\documentclass{article} 

\usepackage{arxiv}
\usepackage[margin=1.48in]{geometry}
\usepackage{natbib}
\usepackage[parfill]{parskip}
\usepackage{hyperref}
\usepackage{url}
\usepackage{enumitem}
\usepackage{float}
\usepackage{algorithm}
\usepackage{algorithmic}

\usepackage{subfigure}
\usepackage{wrapfig}
\usepackage{hyperref}
\usepackage{url}

\usepackage[utf8]{inputenc} 
\usepackage[T1]{fontenc}    
\usepackage{url}            
\usepackage{booktabs}       
\usepackage{amsfonts}       
\usepackage{nicefrac}       
\usepackage{microtype}      
\usepackage{amsmath,amsfonts,graphicx,amssymb,amsthm}
\usepackage{graphicx}  
\usepackage{amsmath,amsfonts}
\usepackage{comment}
\usepackage{titlesec}
\usepackage[capitalise]{cleveref}
\usepackage{algorithm,algorithmic}
\usepackage{enumitem}
\usepackage{caption}

\newcommand\E{\mathbb{E}}

\newcommand{\EE}[2]{\mathbb{E}_{#1}\left[#2\right]}

\newcommand{\addeq}{\addtocounter{equation}{1}\tag{\theequation}}

\newcommand{\D}{\mathcal{D}}
\DeclareMathOperator*{\argmax}{argmax}

\newcommand{\MMD}{\mathrm{MMD}}
\def\pistar{\pi^*}

\newcommand{\algname}{\text{BRAC}}

\newlength{\figwidth}
\title{Behavior Regularized Offline Reinforcement Learning}

\date{}

\author{
Yifan Wu\thanks{Work performed while an intern at Google Brain.}
\\
Carnegie Mellon University\\
\texttt{yw4@cs.cmu.edu}\\
 \And
George Tucker \\
Google Research\\
\texttt{gjt@google.com} \\
 \And
Ofir Nachum \\
Google Research\\
\texttt{ofirnachum@google.com} 
}

\begin{document}

\maketitle

\begin{abstract}
In reinforcement learning (RL) research, it is common to assume access to direct {\em online} interactions with the environment. However in many real-world applications, access to the environment is limited to a fixed {\em offline} dataset of logged experience.
In such settings, standard RL algorithms have been shown to diverge or otherwise yield poor performance. 
Accordingly, recent work has suggested a number of remedies to these issues. 
In this work, we introduce a general framework, \emph{behavior regularized actor critic} ($\algname$), to empirically evaluate recently proposed methods 
as well as a number of simple baselines across a variety of offline continuous control tasks.
Surprisingly, we find that many of the technical complexities introduced in recent methods are unnecessary to achieve  
strong performance.
Additional ablations provide insights into which design choices matter most in the offline RL setting.\footnote{Code is available at \url{https://github.com/google-research/google-research/tree/master/behavior_regularized_offline_rl}.}
\end{abstract}

\section{Introduction}

Offline reinforcement learning (RL) describes the setting in which a learner has access to only a fixed dataset of experience.
In contrast to online RL, additional interactions with the environment during learning are not permitted.
This setting is of particular interest for applications in which deploying a policy is costly 
or there is a safety concern with updating the policy online~\citep{li2015toward}.
For example, for recommendation systems~\citep{li2011unbiased,covington2016deep} or health applications~\citep{murphy2001marginal}, deploying a new policy may only be done at a low frequency after extensive testing and evaluation.
In these cases, the offline dataset is often very large, potentially encompassing years of logged experience. Nevertheless, the inability to interact with the environment directly poses a challenge to modern RL algorithms.

Issues with RL algorithms in the offline setting typically arise in cases where state and actions spaces are large or continuous, necessitating the use of function approximation.
While off-policy (deep) RL algorithms such as DQN~\citep{mnih2013playing}, DDPG~\citep{lillicrap2015continuous}, and SAC~\citep{haarnoja2018soft} may be run directly on offline datasets to learn a policy,
the performance of these algorithms 
has been shown to be sensitive to the experience dataset distribution, even in the online setting when using a replay buffer \citep{van2018deep, fu2019diagnosing}. 
Moreover, \cite{fujimoto2018off} and \cite{kumar2019stabilizing} empirically confirm that
in the offline setting, DDPG fails to learn a good policy, even when the dataset is collected by a single behavior policy, with or without noise added to the behavior policy. 
These failure cases are hypothesized to be caused by erroneous generalization of the state-action value function (Q-value function) learned with function approximators, as suggested by \cite{sutton1995virtues, baird1995residual, tsitsiklis1997analysis, van2018deep}. To remedy this issue, two types of approaches have been proposed recently: 
1) \cite{agarwal2019striving} proposes to apply a random ensemble of Q-value targets to stabilize the learned Q-function, 2)
\cite{fujimoto2018off, kumar2019stabilizing, jaques2019way, laroche2017safe} propose to regularize the learned policy towards the behavior policy based on the intuition that unseen state-action pairs are more likely to receive overestimated Q-values.
These proposed remedies have been shown to improve upon DQN or DDPG at performing policy improvement based on offline data.
Still, each proposal makes several modifications to the building components of baseline off-policy RL algorithms, and each modification may be implemented in various ways.
So a natural question to ask is, which of the design choices in these offline RL algorithms are necessary to achieve good performance? 
For example, to estimate the target Q-value when minimizing the Bellman error, 
\cite{fujimoto2018off} uses a soft combination of two target Q-values, which is different from TD3 \citep{fujimoto2018addressing}, 
where the minimum of two target Q-values is used. 
This soft combination is maintained by \cite{kumar2019stabilizing}, while further increasing the number of Q-networks from two to four. 
As another example, when regularizing towards the behavior policy,
\cite{jaques2019way} uses Kullback-Leibler (KL) divergence with a fixed regularization weight while \cite{kumar2019stabilizing} proposes to use Maximum Mean Discrepancy (MMD) 
with an adaptively trained regularization weight.
Are these design choices crucial to success in offline settings? Or are they simply the result of multiple, human-directed iterations of research?

In this work, we aim at evaluating the importance of different algorithmic building components as well as comparing different design choices in offline RL approaches. 
 We focus on behavior regularized approaches applied to continuous action domains, encompassing many of the recently demonstrated successes~\citep{fujimoto2018off, kumar2019stabilizing}. 
We introduce \emph{behavior regularized actor critic} ($\algname$), 
a general algorithmic framework which covers existing approaches while enabling us to compare the performance of different variants in a modular way. 
We find that many simple variants of the behavior regularized approach can yield good performance, while previously suggested sophisticated techniques such as weighted Q-ensembles and adaptive regularization weights are not crucial.
Experimental ablations reveal further insights into how different design choices affect the performance and robustness of the behavior regularized approach in the offline RL setting. \vspace{-5pt}
\section{Background}

\subsection{Markov Decision Processes}

We consider the standard fully-observed Markov Decision Process (MDP) setting \citep{puterman1990markov}. 
An MDP can be represented as $\mathcal{M} = \left( \mathcal{S}, \mathcal{A}, P, R, \gamma \right)$ where $\mathcal{S}$ is the state space, $\mathcal{A}$ is the action space, $P(\cdot|s, a)$ is the transition probability distribution function, $R(s, a)$ is the reward function and $\gamma$ is the discount factor. 
The goal is to find a policy $\pi(\cdot|s)$ that maximizes the cumulative discounted reward starting from any state $s\in \mathcal{S}$.
Let $P^\pi(\cdot|s)$ denote the induced transition distribution for policy $\pi$.
For later convenience, we also introduce the notion of multi-step transition distributions as $P^\pi_t$,
where $P^\pi_t(\cdot|s)$ denotes the distribution over the state space after rolling out $P^\pi$ for $t$ steps starting from state $s$. 
For example, $P^\pi_0(\cdot|s)$ is the Dirac delta function at $s$ and $P^\pi_1(\cdot|s)=P^\pi(\cdot|s)$. 
We use $R^\pi(s)$ to denote the expected reward at state $s$ when following policy $\pi$, i.e. $R^\pi(s)=\EE{a\sim \pi(\cdot|s)}{R(s,a)}$.  
The state value function (a.k.a. value function) is defined by
$
V^\pi(s) = \sum\nolimits_{t=0}^\infty \gamma^t \EE{s_t\sim P^\pi_t(s)}{R^\pi (s_t)}
$.
The action-value function (a.k.a. Q-function) can be written as
$Q^\pi(s,a) = R(s, a) + \gamma \EE{s'\sim P(\cdot|s, a)}{V^\pi(s')} $.
The optimal policy is defined as the policy $\pistar$ that maximizes $V^{\pistar}(s)$ at all states $s\in \mathcal{S}$. 
In the commonly used actor critic paradigm, one optimizes a policy $\pi_\theta(\cdot|s)$ by alternatively learning a Q-value function $Q_\psi$ to minimize Bellman errors over single step transitions $(s,a,r,s')$,
$
\EE{a'\sim\pi_\theta(\cdot|s')}{\left(r + \gamma \bar{Q}(s',a') - Q_\psi(s,a)\right)^2}
$,
where $\bar{Q}$ denotes a target Q function; e.g., it is common to use a slowly-updated target parameter set $\psi'$ to determine the target Q function as $Q_{\psi'}(s',a')$.
Then, the policy is updated to maximize the Q-values, $\EE{a\sim\pi(\cdot|s)}{Q_\psi(s,a)}$.

\subsection{Offline Reinforcement Learning}

Offline RL (also known as batch RL~\citep{lange2012batch}) considers the problem of learning a policy $\pi$ from a fixed dataset $\D$
consisting of single-step transitions $(s, a, r, s')$. 
Slightly abusing the notion of ``behavior'', we define the behavior policy $\pi_b(a|s)$ as the conditional distribution $p(a|s)$ observed in the dataset distribution $\D$. 
Under this definition, such a behavior policy  $\pi_b$ is always well-defined even if the dataset was collected by multiple, distinct behavior policies.
Because we do not assume direct access to $\pi_b$, it is common in previous work to approximate this behavior policy with max-likelihood over $\mathcal{D}$:
\begin{equation}
    \label{eq:pib}
    \hat{\pi}_b := \argmax_{\hat{\pi}} \EE{(s,a,r,s')\sim\mathcal{D}}{\log \hat{\pi}(a|s)}.
\end{equation}
We denote the learned policy as $\hat{\pi}_b$ and refer to it as the ``cloned policy'' to distinguish it from the true behavior policy.

In this work, we focus on the offline RL problem for complex continuous domains.
We briefly review two recently proposed approaches, BEAR~\citep{kumar2019stabilizing} and BCQ~\citep{fujimoto2018off}.

\paragraph{BEAR}
Motivated by the hypothesis that deep RL algorithms generalize poorly to actions outside the support of the behavior policy, \cite{kumar2019stabilizing} propose BEAR, which learns a policy to maximize Q-values while penalizing it from diverging from behavior policy support. BEAR measures divergence from the behavior policy using kernel MMD~\citep{gretton2007kernel}: 
\begin{align}
\label{eq:kernel-mmd}
\hspace{-0.1in}\MMD^2_k(\pi(\cdot|s), \pi_b(\cdot|s)) = \hspace{-0.1in}\mathop{\E}_{x, x'\sim \pi(\cdot|s)}\left[K(x, x')\right] - 2 \EE{\substack{x\sim \pi(\cdot|s) \\ y\sim \pi_b(\cdot|s)}}{K(x, y)} + \hspace{-0.1in}\mathop{\E}_{y, y'\sim \pi_b(\cdot|s)}\left[K(y, y')\right] \,,
\end{align}
where $K$ is a kernel function. 
Furthermore, to avoid overestimation in the Q-values, the target Q-value function $\bar{Q}$ is calculated as,
\begin{equation}
    \label{eq:bear-q}
    \bar{Q}(s',a'):= 0.75\cdot \min_{j=1,\dots,k} Q_{\psi'_j}(s',a') + 0.25\cdot\max_{j=1,\dots,k} Q_{\psi'_j}(s',a'), 
\end{equation}
where $\psi'_j$ is denotes a soft-updated ensemble of target Q functions. In BEAR's implementation, this ensemble is of size $k=4$. 
BEAR also penalizes target Q-values by an ensemble variance term. However, their empirical results show that there is no clear benefit to doing so, thus we omit this term. 

\paragraph{BCQ}
BCQ enforces $\pi$ to be close to $\pi_b$ with a specific parameterization of $\pi$:
\begin{equation}
    \label{eq:bcq-policy}
    \pi_\theta(a | s) := ~\argmax_{a_i + \xi_\theta(s,a_i)} Q_\psi(s,a_i + \xi_\theta(s,a_i)) 
    ~~~\text{for}~~ a_i\sim\pi_b(a | s), ~ i=1,\dots,N,
\end{equation}
where $\xi_\theta$ is a function approximator with bounded ouptput in $[-\Phi, \Phi]$ where $\Phi$ is a hyperparameter. $N$ is an additional hyperparameter used during evaluation to compute $\pi_\theta$ and during training for Q-value updates.
The target Q-value function $\bar{Q}$ is calculated as in Equation~\ref{eq:bear-q} but with $k=2$.


\section{Behavior Regularized Actor Critic}
Encouraging the learned policy to be close to the behavior policy is a common theme in previous approaches to offline RL. 
To evaluate the effect of different behavior policy regularizers, we introduce 
\emph{behavior regularized actor critic} ($\algname$), an algorithmic framework which generalizes existing approaches while providing more implementation options. 

There are two common ways to incorporate regularization to a specific policy: through a penalty in the value function or as a penalty solely on the policy. 
We begin by introducing the former, \textbf{value penalty (vp)}.  
Similar to SAC~\citep{haarnoja2018soft} which adds an entropy term to the target Q-value calculation, we add a term to the target Q-value calculation that regularizes the learned policy $\pi$ towards the behavior policy $\pi_b$. 
Specifically, we define the penalized value function as
\begin{align*}
V^\pi_D(s) = \sum\nolimits_{t=0}^\infty \gamma^t \EE{s_t\sim P^\pi_t(s)}{R^\pi (s_t) - \alpha D\left(\pi(\cdot|s_t), \pi_b(\cdot|s_t)\right)}\,, \addeq\label{eq:value-penalty}
\end{align*}
where $D$ is a divergence function between distributions over actions (e.g., MMD or KL divergence). 
Following the typical actor critic framework, the Q-value objective is given by,
\begin{align*}
\min_{Q_\psi} ~\EE{\substack{(s,a,r,s')\sim\mathcal{D}\\ a'\sim\pi_\theta(\cdot|s')}}{\left( r + \gamma\left(\bar{Q}(s', a') - \alpha \hat{D}\left(\pi_\theta(\cdot|s'), \pi_b(\cdot|s')\right) \right) - Q_\psi(s, a) \right)^2} \,, \addeq\label{eq:q-obj}
\end{align*}
where $\bar{Q}$ again denotes a target Q function 
and $\hat{D}$ denotes a sample-based estimate of the divergence function $D$. The policy learning objective can be written as,
\begin{align*}
\max_{\pi_\theta} ~\EE{(s,a,r,s')\sim\mathcal{D}}{ ~ \EE{a''\sim \pi_\theta(\cdot|s)}{Q_\psi(s, a'')} - \alpha \hat{D}\left(\pi_\theta(\cdot|s), \pi_b(\cdot|s)\right) } \,. \addeq\label{eq:pi-obj}
\end{align*}
Accordingly, one performs alternating gradient updates based on \eqref{eq:q-obj} and \eqref{eq:pi-obj}. 
This algorithm is equivalent to SAC when using a single-sample estimate of the entropy for $\hat{D}$; i.e., $\hat{D}(\pi_\theta(\cdot|s'),\pi_b(\cdot|s')):= \log \pi(a'|s')$ for $a'\sim\pi(\cdot|s')$.

The second way to add the regularizer is to only regularize the policy during policy optimization.
That is, we use the same objectives in Equations~\ref{eq:q-obj} and~\ref{eq:pi-obj}, but use $\alpha=0$ in the Q update while using a non-zero $\alpha$ in the policy update.
We call this variant \textbf{policy regularization (pr)}. 
This proposal is similar to the regularization employed in A3C~\citep{mnih2016asynchronous}, if one uses the entropy of $\pi_\theta$ to compute $\hat{D}$.

In addition to the choice of value penalty or policy regularization, the choice of $D$ and how to perform sample estimation of $\hat{D}$ is a key design choice of $\algname$:

\paragraph{Kernel MMD}
We can compute a sample based estimate of kernel MMD (Equation~\ref{eq:kernel-mmd}) by drawing samples from both $\pi_\theta$ and $\pi_b$. Because we do not have access to multiple samples from $\pi_b$, this requires a pre-estimated cloned policy $\hat{\pi}_b$.

\paragraph{KL Divergence}
With KL Divergence, the behavior regularizer can be written as
\begin{align*}
D_{\mathrm{KL}} \left(\pi_\theta(\cdot|s), \pi_b(\cdot|s)\right) 
= \EE{a\sim \pi_\theta(\cdot|s)}{\log\pi_\theta(a|s) - \log \pi_b(a|s)}
\,.
\end{align*}
Directly estimating $D_{\mathrm{KL}}$ via samples requires having access to the density of both $\pi_\theta$ and $\pi_b$; as in MMD, the cloned $\hat{\pi}_b$ can be used in place of $\pi_b$. 
Alternatively, we can avoid estimating $\pi_b$ explicitly, 
by using the dual form of the KL-divergence. Specifically, any $f$-divergence \citep{csiszar1964informationstheoretische} has a dual form~\citep{nowozin2016f} given by, 
\begin{align*}
D_f(p, q) = \EE{x\sim p}{f\left(q(x)/p(x)\right)} = \max_{g:\mathcal{X}\mapsto \mathrm{dom}(f^*)}
\EE{x\sim q}{g(x)} - \EE{x\sim p}{f^*(g(x))} \,,
\end{align*}
where $f^*$ is the Fenchel dual of $f$. 
In this case, one no longer needs to estimate a cloned policy $\hat{\pi}_b$ but instead needs to learn a discriminator function $g$ with minimax optimization as in \cite{nowozin2016f}. This sample based dual estimation can be applied to any $f$-divergence. 
In the case of a KL-divergence, $f(x)=- \log x$ and $f^*(t)=-\log(-t)-1$.

\paragraph{Wasserstein Distance}  One may also use the Wassertein distance as the divergence $D$. For sample-based estimation, one may use its dual form,
\begin{align*}
W(p, q) = \sup_{g:||g||_L\le 1} \EE{x\sim p}{g(x)} - \EE{x\sim q}{g(x)}
\end{align*}
and maintain a discriminator $g$ as in \cite{gulrajani2017improved}.

Now we discuss how existing approaches can be instantiated under the framework of $\algname$.

\paragraph{BEAR}
To re-create BEAR with $\algname$, one uses policy regularization with the sample-based kernel MMD for $\hat{D}$ and uses a min-max ensemble estimate for $\bar{Q}$ (Equation~\ref{eq:bear-q}).
Furthermore, BEAR adaptively trains the regularization weight $\alpha$ as a Lagriagian multiplier: it sets a threshold $\epsilon>0$ for the kernel MMD distance and increases $\alpha$ if the current average divergence is above the threshold and decreases $\alpha$ if below the threshold. 

\paragraph{BCQ}
The BCQ algorithm does not use any regularizers (i.e. $\alpha=0$ for both value and policy objectives).
Still, the algorithm may be realized by $\algname$ if one restricts the policy optimization in Equation~\ref{eq:pi-obj} to be over parameterized policies based on Equation~\ref{eq:bcq-policy}.


\paragraph{KL-Control}
There has been a rich set of work which investigates regularizing the learned policy through KL-divergence with respect to another policy, e.g. 
\cite{abdolmaleki2018maximum, kakade2002natural, peters2010relative, schulman2015trust, nachum2017trust}. Notably, \cite{jaques2019way} apply this idea to offline RL in discrete action domains by introducing a KL value penalty in the Q-value definition. 
It is clear that $\algname$ can realize this algorithm as well.

To summarize, one can instantiate the behavior regularized actor critic framework with different design choices, 
including how to estimate the target Q value, 
which divergence to use, 
whether to learn $\alpha$ adaptively, 
whether to use a value penalty in the Q function objective \eqref{eq:q-obj} or just use policy regularization in \eqref{eq:pi-obj} and so on. 
In the next section, we empirically evaluate a set of these different design choices to provide insights into what actually matters when approaching the offline RL problem.




\section{Experiments}
The $\algname$ framework encompasses several previously proposed methods depending on specific design choices (e.g., whether to use value penalty or policy regularization, how to compute the target Q-value, and how to impose the behavior regularization).
For a practitioner, key questions are: How should these design choices be made?  Which variations among these different algorithms actually matter?
To answer these questions,  
we perform a systematic evaluation of $\algname$ under different design choices.

Following \cite{kumar2019stabilizing}, we evaluate performance on four Mujoco \citep{todorov2012mujoco} continuous control environments in OpenAI Gym \citep{brockman2016openai}: Ant-v2, HalfCheetah-v2, Hopper-v2, and Walker2d-v2. In many real-world applications of RL, one has logged data from sub-optimal policies (e.g., robotic control and recommendation systems). To simulate this scenario, we collect the offline dataset with a sub-optimal policy perturbed by additional noise. 
To obtain a partially trained policy, we train a policy with SAC and online interactions until the policy performance achieves a performance threshold ($1000, 4000, 1000, 1000$ for Ant-v2, HalfCheetah-v2, Hopper-v2, Walker2d-v2, respectively, similar to the protocol established by~\citet{kumar2019stabilizing}). 
Then, we perturb the partially trained policy with noise (Gaussian noise or $\epsilon$-greedy at different levels) to simulate different exploration strategies resulting in five noisy behavior policies. We collect 1 million transitions according to each behavior policy resulting in five datasets for each environment (see Appendix for implementation details).
We evaluate offline RL algorithms by training on these fixed datasets and evaluating the learned policies on the real environments. 

In preliminary experiments, we found that policy learning rate and regularization strength have a significant effect on performance. 
As a result, for each variant of $\algname$ and each environment, we do a grid search over policy learning rate and regularization strength. 
For policy learning rate, we search over six values, ranging from $3\cdot 10^6$ to $0.001$. 
The regularization strength is controlled differently in different algorithms. In the simplest case, the regularization weight $\alpha$ is fixed; in BEAR the regularization weight is adaptively trained with dual gradient ascent based on a divergence constraint $\epsilon$ that is tuned as a hyperparameter;
in BCQ the corresponding tuning is for the perturbation range $\Phi$.
For each of these options, we search over five values (see Appendix for details).
For existing algorithms such as BEAR and BCQ, the reported hyperparameters in their papers \citep{kumar2019stabilizing, fujimoto2018off} are included in this search range,
We select the best hyperparameters according to the average performance over all five datasets. 

Currently, BEAR~\citep{kumar2019stabilizing} provides state-of-the-art performance on these tasks, so to understand the effect of variations under our $\algname$ framework, we start by implementing BEAR in $\algname$ and run a series of comparisons by varying different design choices: adaptive vs. fixed regularization, different ensembles for estimating target Q-values, value penalty vs. policy regularization and divergence choice for the regularizer.
We then evaluate BCQ, which has a different design in the $\algname$ framework, and compare it to other $\algname$ variants as well as several baseline algorithms.

\subsection{Fixed v.s. adaptive regularization weights}

\begin{figure}[h]
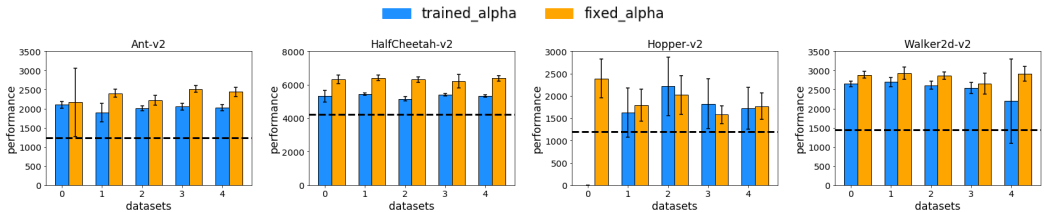

\vspace{-.15in}
\centering
\includegraphics[height=0.035\columnwidth]{figs_2/fixed_alpha_bar_legend.pdf}\\
\includegraphics[width=0.24\columnwidth]{figs_2/fixed_alpha_bar_Ant-v2.pdf}
\includegraphics[width=0.24\columnwidth]{figs_2/fixed_alpha_bar_HalfCheetah-v2.pdf}
\includegraphics[width=0.24\columnwidth]{figs_2/fixed_alpha_bar_Hopper-v2.pdf}
\includegraphics[width=0.24\columnwidth]{figs_2/fixed_alpha_bar_Walker2d-v2.pdf}
\caption{
Comparing fixed $\alpha$ with adaptively trained $\alpha$. Black dashed lines are the performance of the partially trained policies (distinct from the behavior policies which have injected noise). 
We report the mean over the last 10 evaluation points (during training) averaged over 5 different random seeds. Each evaluation point is the return averaged over 20 episodes. Error bars represent the standard deviation across different random seeds.
We report the performance as 0 if it is negative.
}
\label{fig:fix_alpha}
\vspace{-0.15in}
\end{figure}

In BEAR, regularization is controlled by a threshold $\epsilon$, which is used for adaptively training the Lagrangian multiplier $\alpha$, whereas typically (e.g., in KL-control) one uses a fixed $\alpha$. In our initial experiments with BEAR, we found that when using the recommended value of $\epsilon$, the learned value of $\alpha$ consistently increased during training, implying that the MMD constraint between $\pi_\theta$ and $\pi_b$ was almost never satisfied. This suggests that BEAR is effectively performing policy regularization with a large $\alpha$ rather than constrained optimization.
This led us to question if adaptively training $\alpha$ is better than using a fixed $\alpha$.
To investigate this question, we evaluate the performance of both approaches (with appropriate hyperparameter tuning for each, over either $\alpha$ or $\epsilon$) in Figure~\ref{fig:fix_alpha}.
On most datasets, both approaches learn a policy that is much better than the \emph{partially trained policy}\footnote{The partially trained policy is the policy used to collect data \emph{without} injected noise. The true \emph{behavior policy} and \emph{behavior cloning} will usually get worse performance due to injected noise when collecting the data.},
although we do observe a consistent modest advantage when using a fixed $\alpha$.
Because using a fixed $\alpha$ is simpler and performs better than adaptive training, we use this approach in subsequent experiments.

\subsection{Ensemble for target Q-values}

\begin{figure}[h]
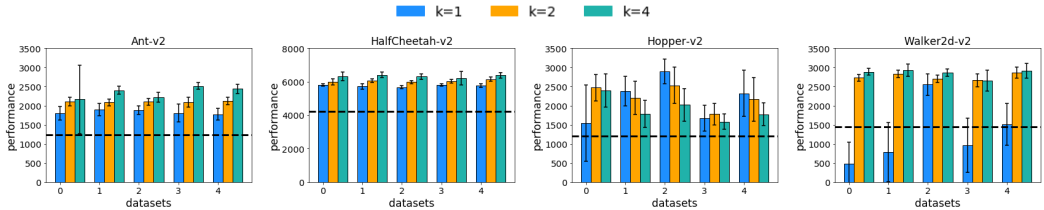

\vspace{-.15in}
\centering
\includegraphics[height=0.035\columnwidth]{figs_2/ens_k_bar_legend.pdf}\\
\includegraphics[width=0.24\columnwidth]{figs_2/ens_k_bar_Ant-v2.pdf}
\includegraphics[width=0.24\columnwidth]{figs_2/ens_k_bar_HalfCheetah-v2.pdf}
\includegraphics[width=0.24\columnwidth]{figs_2/ens_k_bar_Hopper-v2.pdf}
\includegraphics[width=0.24\columnwidth]{figs_2/ens_k_bar_Walker2d-v2.pdf}
\caption{
Comparing different number of Q-functions for target Q-value ensemble. We use a weighted mixture to compute the target value for all of these variants. As expected, we find that using an ensemble ($k>1$) is better than using a single Q-function. 
}
\label{fig:ensemble_k}
\vspace{-0.15in}
\end{figure}

Another important design choice in $\algname$ is how to compute the target Q-value, and specifically, whether one should use the sophisticated ensemble strategies employed by BEAR and BCQ.
Both BEAR and BCQ use a weighted mixture of the minimum and maximum among multiple learned Q-functions (compared to TD3 which simply uses the minimum of two). BEAR further increases the number of Q-functions from 2 to 4. To investigate these design choices, we first experiment with different number of Q-functions $k=\{1, 2, 4\}$. Results are shown in Figure~\ref{fig:ensemble_k}. \citet{fujimoto2018addressing} show that using two Q-functions provides significant improvements in online RL; similarly, we find that using $k=1$ sometimes fails to learn a good policy (e.g., in Walker2d) in the offline setting. Using $k=4$ has a small advantage compared to $k=2$ except in Hopper. Both $k=2$ and $k=4$ significantly improve over the partially trained policy baseline. In general, increasing the value of $k$ in ensemble will lead to more stable or better performance, but requires more computation cost. On these domains we found that $k=4$ only gives marginal improvement over $k=2$, so we use $k=2$ in our remaining experiments. 

\begin{figure}[h]
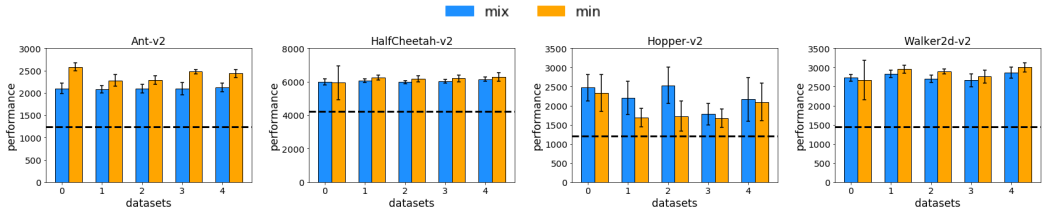

\centering
\includegraphics[height=0.035\columnwidth]{figs_2/ens_mix_bar_legend.pdf}\\
\includegraphics[width=0.24\columnwidth]{figs_2/ens_mix_bar_Ant-v2.pdf}
\includegraphics[width=0.24\columnwidth]{figs_2/ens_mix_bar_HalfCheetah-v2.pdf}
\includegraphics[width=0.24\columnwidth]{figs_2/ens_mix_bar_Hopper-v2.pdf}
\includegraphics[width=0.24\columnwidth]{figs_2/ens_mix_bar_Walker2d-v2.pdf}
\caption{
Comparing taking the minimum v.s. a weighted mixture in Q-value ensemble. We find that simply taking the minimum is usually slightly better, except in Hopper-v2.
}
\label{fig:ensemble_mix}
\vspace{-0.15in}
\end{figure}

Regarding whether using a weighed mixture of Q-values or the minimum, we compare these two options under $k=2$. Results are shown in Figure~\ref{fig:ensemble_mix}. We find that taking the minimum performs slightly better than taking a mixture except in Hopper, and both successfully outperform the partially trained policy in all cases. Due to the simplicity and strong performance of taking the minimum of two Q-functions, we use this approach in subsequent experiments.

\subsection{Value penalty or policy regularization}

\begin{figure}[h]
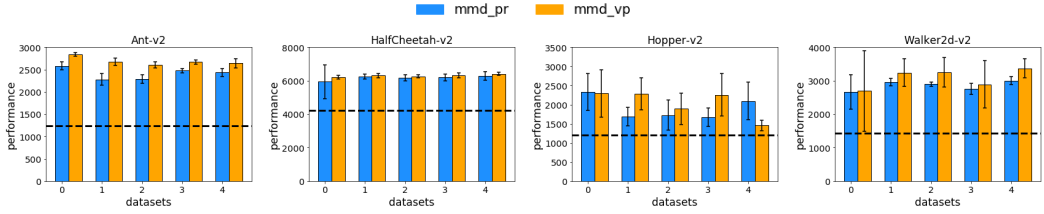

\vspace{-0.15in}
\centering
\includegraphics[height=0.035\columnwidth]{figs_2/pr_vp_mmd_legend.pdf}\\
\includegraphics[width=0.24\columnwidth]{figs_2/pr_vp_mmd_Ant-v2.pdf}
\includegraphics[width=0.24\columnwidth]{figs_2/pr_vp_mmd_HalfCheetah-v2.pdf}
\includegraphics[width=0.24\columnwidth]{figs_2/pr_vp_mmd_Hopper-v2.pdf}
\includegraphics[width=0.24\columnwidth]{figs_2/pr_vp_mmd_Walker2d-v2.pdf}
\caption{
Comparing policy regularization (pr) v.s. value penalty (vp) with MMD. The use of value penalty is usually slightly better.
}
\label{fig:bear_pr_vp}
\vspace{-0.15in}
\end{figure}

So far, we have evaluated variations in regularization weights and ensemble of Q-values. We found that the technical complexity introduced in recent works is not always necessary to achieve state-of-the-art performance. With these simplifications, we now evaluate a major variation of design choices in $\algname$ --- using \emph{value penalty} or \emph{policy regularization}. We follow our simplified version of BEAR: MMD policy regularization, fixed $\alpha$, and computation of target Q-values based on the minimum of a $k=2$ ensemble. We compare this instantiation of $\algname$ to its value penalty version, with results shown in Figure~\ref{fig:bear_pr_vp}. While both variants outperform the partially trained policy, we find that value penalty performs slightly better than policy regularization in most cases. We consistently observed this advantage with other divergence choices (see Appendix Figure~\ref{fig:pr_vp_all} for a full comparison).  

\subsection{Divergences for regularization}

\begin{figure}[h]
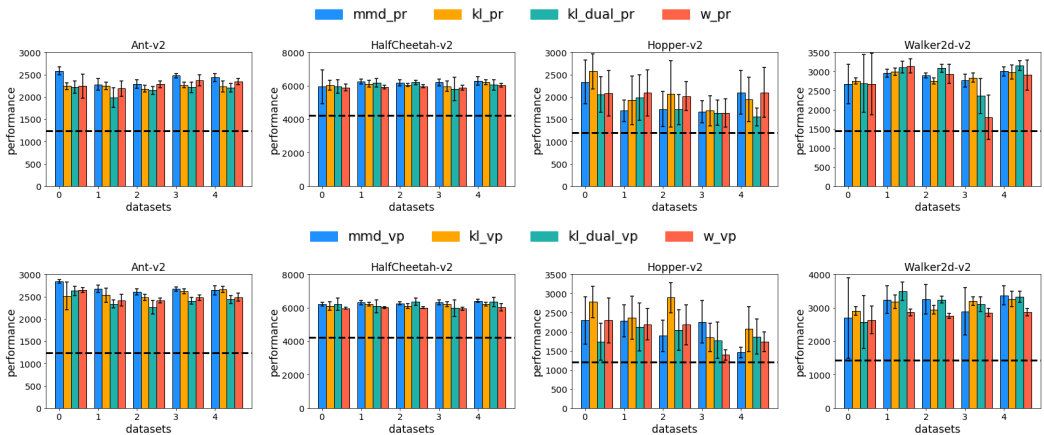

\vspace{-0.15in}
\centering
\includegraphics[height=0.035\columnwidth]{figs_2/div_pr_legend.pdf}\\
\includegraphics[width=0.24\columnwidth]{figs_2/div_pr_Ant-v2.pdf}
\includegraphics[width=0.24\columnwidth]{figs_2/div_pr_HalfCheetah-v2.pdf}
\includegraphics[width=0.24\columnwidth]{figs_2/div_pr_Hopper-v2.pdf}
\includegraphics[width=0.24\columnwidth]{figs_2/div_pr_Walker2d-v2.pdf}\\
\includegraphics[height=0.035\columnwidth]{figs_2/div_vp_legend.pdf}\\
\includegraphics[width=0.24\columnwidth]{figs_2/div_vp_Ant-v2.pdf}
\includegraphics[width=0.24\columnwidth]{figs_2/div_vp_HalfCheetah-v2.pdf}
\includegraphics[width=0.24\columnwidth]{figs_2/div_vp_Hopper-v2.pdf}
\includegraphics[width=0.24\columnwidth]{figs_2/div_vp_Walker2d-v2.pdf}
\caption{
Comparing different divergences under both policy regularization (top row) and value penalty (bottom row). All variants yield similar performance, which is significantly better than the partially trained policy.
}
\label{fig:div}
\vspace{-0.15in}
\end{figure}

We evaluated four choices of divergences used as the regularizer $D$: (a) MMD (as in BEAR), (b) KL in the primal form with estimated behavior policy (as in KL-control), and (c) KL and (d) Wasserstein in their dual forms without estimating a behavior policy. As shown in Figure~\ref{fig:div}, we do not find any specific divergence performing consistently better or worse than the others. All variants are able to learn a policy that significantly improves over the behavior policy in all cases. 

In contrast, \citet{kumar2019stabilizing} argue that sampled MMD is superior to KL based on the idea that
it is better to regularize the support of the learned policy distribution to be within the support of the behavior policy rather than forcing the two distributions to be similar. 
While conceptually reasonable, we do not find support for that argument in our experiments: (i) we find that KL and Wassertein can perform similarly well to MMD even though they are not designed for support matching; (ii) we briefly tried  divergences that are explicitly designed for support matching (the relaxed KL and relaxed Wasserstein distances proposed by \cite{wu2019domain}), but did not observe a clear benefit to the additional complexity. We conjecture that this is because even if one uses noisy or multiple behavior policies to collect data, the noise is reflected more in the diversity of states rather than the diversity of actions on a single state (due to the nature of environment dynamics). However, we expect this support matching vs. distribution matching distinction may matter in other scenarios such as smaller state spaces or contextual bandits, which is a potential direction for future work.

\subsection{Comparison to BCQ and other baselines}

\begin{figure}[h]
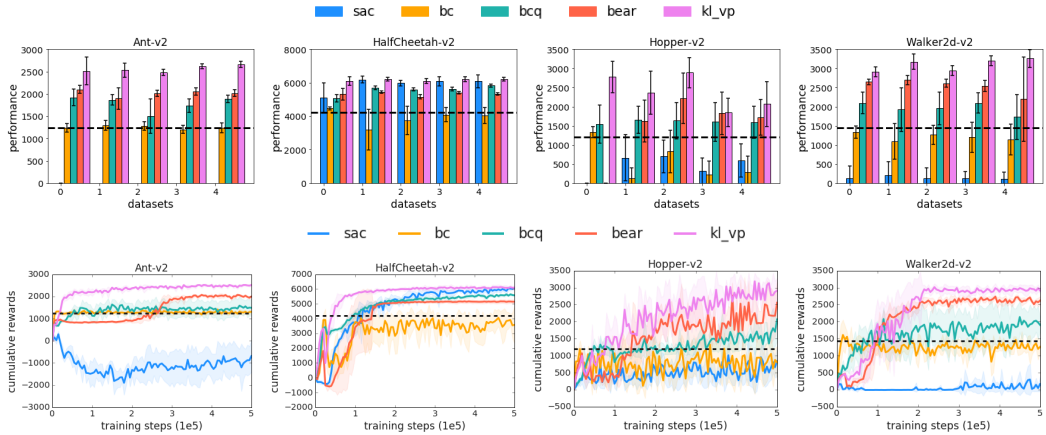

\vspace{-0.15in}
\centering
\includegraphics[height=0.035\columnwidth]{figs_2/baselines_legend.pdf}\\
\includegraphics[width=0.24\columnwidth]{figs_2/baselines_Ant-v2.pdf}
\includegraphics[width=0.24\columnwidth]{figs_2/baselines_HalfCheetah-v2.pdf}
\includegraphics[width=0.24\columnwidth]{figs_2/baselines_Hopper-v2.pdf}
\includegraphics[width=0.24\columnwidth]{figs_2/baselines_Walker2d-v2.pdf}
\includegraphics[height=0.035\columnwidth]{figs_curves/bcq_legend.pdf}\\
\includegraphics[width=0.24\columnwidth]{figs_curves/bcq_Ant-v2_eps3.pdf}
\includegraphics[width=0.24\columnwidth]{figs_curves/bcq_HalfCheetah-v2_eps3.pdf}
\includegraphics[width=0.24\columnwidth]{figs_curves/bcq_Hopper-v2_eps3.pdf}
\includegraphics[width=0.24\columnwidth]{figs_curves/bcq_Walker2d-v2_eps3.pdf}
\caption{
Comparing value penalty with KL divergence (kl\_vp) to vanilla SAC, behavior cloning (bc), BCQ and BEAR. Bottom row shows sampled training curves with 1 out of the 5 datasets. See Appendix for training curves on all datasets.
}
\label{fig:bcq}
\vspace{-0.15in}
\end{figure}

We now compare one of our best performing algorithms so far, kl\_vp (value penalty with KL divergence in the primal form), to BCQ, BEAR, and two other baselines: vanilla SAC (which uses adaptive entropy regularization) and behavior cloning. Figure~\ref{fig:bcq} shows the comparison. We find that vanilla SAC only works in the HalfCheetah environment and fails in the other three environments. Behavior cloning never learns a better policy than the partially trained policy used to collect the data. Although BCQ consistently learns a policy that is better than the partially trained policy, its performance is always clearly worse than kl\_vp (and other variants whose performance is similar to kl\_vp, according to our previous experiments). We conclude that BCQ is less favorable than explicitly using a divergence for behavior regularization (BEAR and kl\_vp). Although, tuning additional hyperparameters beyond $\Phi$ for  BCQ may improve performance.

\subsection{Hyperparameter Sensitivity} 

In our experiments, we find that many simple algorithmic designs  achieve good performance under the framework of $\algname$. For example, all of the 4 divergences we tried perform similarly well when used for regularization. 
In these experiments, we allowed for appropriate hyperparameter tuning over policy learning rate and regularization weight, as we initially found that not doing so can lead to premature and incorrect conclusions.
\footnote{
For example, taking the optimal hyperparameters from one design choice and then applying them to a different design choice (e.g., MMD vs KL divergence) can lead to incorrect conclusions (specifically, that using KL is worse than using MMD, only because one transferred the hyperparameters used for MMD to KL).
}
However, some design choices may be more robust to hyperparameters than others. 
To investigate this, 
we also analyzed the sensitivity to hyperparameters for all algorithmic variants (Appendix Figures~\ref{fig:grid-search-full} and~\ref{fig:grid-search-baselines}). 
To summarize, we found that 
(i) MMD and KL Divergence are similar in terms of sensitivity to hyperparameters,
(ii) using the dual form of divergences (e.g. KL dual, Wasserstein) appears to be more sensitive to hyperparameters, possibly because of the more complex training procedure (optimizing a minimax objective), and 
(iii) value penalty is slightly more sensitive to hyperparameters than policy regularization despite its more favorable performance under the best hyperparameters. 

\begin{figure}[h]
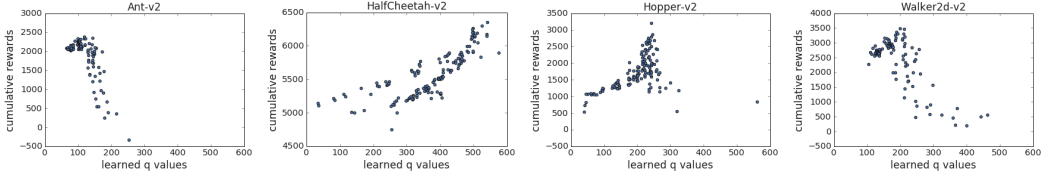

\centering
\includegraphics[width=0.24\columnwidth]{figs_corr/Ant-v2_mmd_pr_eps1.pdf}
\includegraphics[width=0.24\columnwidth]{figs_corr/HalfCheetah-v2_kl_pr_eps1.pdf}
\includegraphics[width=0.24\columnwidth]{figs_corr/Hopper-v2_kl_vp_eps1.pdf}
\includegraphics[width=0.24\columnwidth]{figs_corr/Walker2d-v2_mmd_pr_eps1.pdf}
\caption{
Correlation between learned Q-values and performance. x-axis is the average of learned $Q_\psi(s,a)$ over the last 500 training batches. y-axis is the average performance over the last 10 evaluation points. Each plot corresponds to a (environment, algorithm, dataset) tuple. Different points in each plot correspond to different hyperparameters and different random seeds. 
}
\label{fig:corr}
\end{figure}

Although we utilized hyperparameter searches in our results, in pure offline RL settings, testing on the real environment is infeasible. Thus, a natural question is how to select the best hyperparameter or the best learned policy among many without direct testing.
As a preliminary attempt, we evaluated whether the Q-values learned during training can be used as a proxy for hyperparameter selection. 
Specifically, we look at the correlation between the average learned Q-values (in mini-batches) and the true performance. Figure~\ref{fig:corr} shows sampled visualizations of these Q-values. 
We find that the learned Q-values are not a good indicator of the performance, even when they are within a reasonable range (i.e., not diverging during training). 
A more formal direction for doing hyperparameter selection is to do off-policy evaluation.
However, off-policy evaluation is an open research problem with limited success on complex continuous control tasks (see \citet{liu2018breaking, nachum2019dualdice, irpan2019off} for recent attempts), we leave hyperparameter selection as future work and encourage more researchers to investigate this direction.
\vspace{-0.1in}
\section{Conclusion}
\vspace{-0.1in}
In this work, we introduced \emph{behavior regularized actor critic} ($\algname$), an algorithmic framework, which generalizes existing approaches to solve the offline RL problem by regularizing to the behavior policy.
In our experiments, we showed that many sophisticated training techniques, such as weighted target Q-value ensembles and adaptive regularization coefficients are not necessary in order to achieve state-of-the-art performance.
We found that the use of value penalty is slightly better than policy regularization, while many possible divergences (KL, MMD, Wasserstein) can achieve similar performance.
Perhaps the most important differentiator in these offline settings is whether proper hyperparameters are used.
Although some variants of $\algname$ are more robust to hyperparameters than others, every variant relies on a suitable set of hyperparameters to train well.
Off-policy evaluation without interacting with the environment is a challenging open problem. While previous off-policy evaluation work focuses on reducing mean-squared-error to the expected return, in our problem, we only require a ranking of policies. This relaxation may allow novel solutions, and 
we encourage more researchers to investigate this direction in the pursuit of truly offline RL.
Another potential direction is to look at the situations when the dataset is much smaller. Our preliminary observations on smaller datasets is that it is hard to get a hyperparameter that works consistently well on multiple runs with different random seeds. So we conjecture that smaller datasets may need either more careful hyperparameter search or (more interestingly) a better algorithm. We leave an extensive study of this setting to future work.

\section*{Acknowledgements}
We thank Aviral Kumar, Ilya Kostrikov, Yinlam Chow, and others at Google Research for helpful thoughts and discussions.

\bibliographystyle{iclr2020_conference}
\bibliography{offlinerl}

\begin{thebibliography}{36}
\providecommand{\natexlab}[1]{#1}
\providecommand{\url}[1]{\texttt{#1}}
\expandafter\ifx\csname urlstyle\endcsname\relax
  \providecommand{\doi}[1]{doi: #1}\else
  \providecommand{\doi}{doi: \begingroup \urlstyle{rm}\Url}\fi

\bibitem[Abdolmaleki et~al.(2018)Abdolmaleki, Springenberg, Tassa, Munos,
  Heess, and Riedmiller]{abdolmaleki2018maximum}
Abbas Abdolmaleki, Jost~Tobias Springenberg, Yuval Tassa, Remi Munos, Nicolas
  Heess, and Martin Riedmiller.
\newblock Maximum a posteriori policy optimisation.
\newblock \emph{arXiv preprint arXiv:1806.06920}, 2018.

\bibitem[Agarwal et~al.(2019)Agarwal, Schuurmans, and
  Norouzi]{agarwal2019striving}
Rishabh Agarwal, Dale Schuurmans, and Mohammad Norouzi.
\newblock Striving for simplicity in off-policy deep reinforcement learning.
\newblock \emph{arXiv preprint arXiv:1907.04543}, 2019.

\bibitem[Baird(1995)]{baird1995residual}
Leemon Baird.
\newblock Residual algorithms: Reinforcement learning with function
  approximation.
\newblock In \emph{Machine Learning Proceedings 1995}, pp.\  30--37. Elsevier,
  1995.

\bibitem[Brockman et~al.(2016)Brockman, Cheung, Pettersson, Schneider,
  Schulman, Tang, and Zaremba]{brockman2016openai}
Greg Brockman, Vicki Cheung, Ludwig Pettersson, Jonas Schneider, John Schulman,
  Jie Tang, and Wojciech Zaremba.
\newblock Openai gym.
\newblock \emph{arXiv preprint arXiv:1606.01540}, 2016.

\bibitem[Covington et~al.(2016)Covington, Adams, and Sargin]{covington2016deep}
Paul Covington, Jay Adams, and Emre Sargin.
\newblock Deep neural networks for youtube recommendations.
\newblock In \emph{Proceedings of the 10th ACM conference on recommender
  systems}, pp.\  191--198. ACM, 2016.

\bibitem[Csisz{\'a}r(1964)]{csiszar1964informationstheoretische}
Imre Csisz{\'a}r.
\newblock Eine informationstheoretische ungleichung und ihre anwendung auf
  beweis der ergodizitaet von markoffschen ketten.
\newblock \emph{Magyer Tud. Akad. Mat. Kutato Int. Koezl.}, 8:\penalty0
  85--108, 1964.

\bibitem[Fu et~al.(2019)Fu, Kumar, Soh, and Levine]{fu2019diagnosing}
Justin Fu, Aviral Kumar, Matthew Soh, and Sergey Levine.
\newblock Diagnosing bottlenecks in deep q-learning algorithms.
\newblock \emph{arXiv preprint arXiv:1902.10250}, 2019.

\bibitem[Fujimoto et~al.(2018{\natexlab{a}})Fujimoto, Meger, and
  Precup]{fujimoto2018off}
Scott Fujimoto, David Meger, and Doina Precup.
\newblock Off-policy deep reinforcement learning without exploration.
\newblock \emph{arXiv preprint arXiv:1812.02900}, 2018{\natexlab{a}}.

\bibitem[Fujimoto et~al.(2018{\natexlab{b}})Fujimoto, van Hoof, and
  Meger]{fujimoto2018addressing}
Scott Fujimoto, Herke van Hoof, and David Meger.
\newblock Addressing function approximation error in actor-critic methods.
\newblock \emph{arXiv preprint arXiv:1802.09477}, 2018{\natexlab{b}}.

\bibitem[Gretton et~al.(2007)Gretton, Borgwardt, Rasch, Sch{\"o}lkopf, and
  Smola]{gretton2007kernel}
Arthur Gretton, Karsten~M Borgwardt, Malte Rasch, Bernhard Sch{\"o}lkopf, and
  Alexander~J Smola.
\newblock A kernel approach to comparing distributions.
\newblock In \emph{Proceedings of the National Conference on Artificial
  Intelligence}, volume~22, pp.\  1637. Menlo Park, CA; Cambridge, MA; London;
  AAAI Press; MIT Press; 1999, 2007.

\bibitem[Gulrajani et~al.(2017)Gulrajani, Ahmed, Arjovsky, Dumoulin, and
  Courville]{gulrajani2017improved}
Ishaan Gulrajani, Faruk Ahmed, Martin Arjovsky, Vincent Dumoulin, and Aaron~C
  Courville.
\newblock Improved training of wasserstein gans.
\newblock In \emph{Advances in neural information processing systems}, pp.\
  5767--5777, 2017.

\bibitem[Haarnoja et~al.(2018)Haarnoja, Zhou, Abbeel, and
  Levine]{haarnoja2018soft}
Tuomas Haarnoja, Aurick Zhou, Pieter Abbeel, and Sergey Levine.
\newblock Soft actor-critic: Off-policy maximum entropy deep reinforcement
  learning with a stochastic actor.
\newblock \emph{arXiv preprint arXiv:1801.01290}, 2018.

\bibitem[Irpan et~al.(2019)Irpan, Rao, Bousmalis, Harris, Ibarz, and
  Levine]{irpan2019off}
Alex Irpan, Kanishka Rao, Konstantinos Bousmalis, Chris Harris, Julian Ibarz,
  and Sergey Levine.
\newblock Off-policy evaluation via off-policy classification.
\newblock \emph{arXiv preprint arXiv:1906.01624}, 2019.

\bibitem[Jaques et~al.(2019)Jaques, Ghandeharioun, Shen, Ferguson, Lapedriza,
  Jones, Gu, and Picard]{jaques2019way}
Natasha Jaques, Asma Ghandeharioun, Judy~Hanwen Shen, Craig Ferguson, Agata
  Lapedriza, Noah Jones, Shixiang Gu, and Rosalind Picard.
\newblock Way off-policy batch deep reinforcement learning of implicit human
  preferences in dialog.
\newblock \emph{arXiv preprint arXiv:1907.00456}, 2019.

\bibitem[Kakade(2002)]{kakade2002natural}
Sham~M Kakade.
\newblock A natural policy gradient.
\newblock In \emph{Advances in neural information processing systems}, pp.\
  1531--1538, 2002.

\bibitem[Kumar et~al.(2019)Kumar, Fu, Tucker, and Levine]{kumar2019stabilizing}
Aviral Kumar, Justin Fu, George Tucker, and Sergey Levine.
\newblock Stabilizing off-policy q-learning via bootstrapping error reduction.
\newblock \emph{arXiv preprint arXiv:1906.00949}, 2019.

\bibitem[Lange et~al.(2012)Lange, Gabel, and Riedmiller]{lange2012batch}
Sascha Lange, Thomas Gabel, and Martin Riedmiller.
\newblock Batch reinforcement learning.
\newblock In \emph{Reinforcement learning}, pp.\  45--73. Springer, 2012.

\bibitem[Laroche \& Trichelair(2017)Laroche and Trichelair]{laroche2017safe}
Romain Laroche and Paul Trichelair.
\newblock Safe policy improvement with baseline bootstrapping.
\newblock \emph{arXiv preprint arXiv:1712.06924}, 2017.

\bibitem[Li et~al.(2011)Li, Chu, Langford, and Wang]{li2011unbiased}
Lihong Li, Wei Chu, John Langford, and Xuanhui Wang.
\newblock Unbiased offline evaluation of contextual-bandit-based news article
  recommendation algorithms.
\newblock In \emph{Proceedings of the fourth ACM international conference on
  Web search and data mining}, pp.\  297--306. ACM, 2011.

\bibitem[Li et~al.(2015)Li, Munos, and Szepesv{\'a}ri]{li2015toward}
Lihong Li, R{\'e}mi Munos, and Csaba Szepesv{\'a}ri.
\newblock Toward minimax off-policy value estimation.
\newblock 2015.

\bibitem[Lillicrap et~al.(2015)Lillicrap, Hunt, Pritzel, Heess, Erez, Tassa,
  Silver, and Wierstra]{lillicrap2015continuous}
Timothy~P Lillicrap, Jonathan~J Hunt, Alexander Pritzel, Nicolas Heess, Tom
  Erez, Yuval Tassa, David Silver, and Daan Wierstra.
\newblock Continuous control with deep reinforcement learning.
\newblock \emph{arXiv preprint arXiv:1509.02971}, 2015.

\bibitem[Liu et~al.(2018)Liu, Li, Tang, and Zhou]{liu2018breaking}
Qiang Liu, Lihong Li, Ziyang Tang, and Dengyong Zhou.
\newblock Breaking the curse of horizon: Infinite-horizon off-policy
  estimation.
\newblock In \emph{Advances in Neural Information Processing Systems}, pp.\
  5356--5366, 2018.

\bibitem[Mnih et~al.(2013)Mnih, Kavukcuoglu, Silver, Graves, Antonoglou,
  Wierstra, and Riedmiller]{mnih2013playing}
Volodymyr Mnih, Koray Kavukcuoglu, David Silver, Alex Graves, Ioannis
  Antonoglou, Daan Wierstra, and Martin Riedmiller.
\newblock Playing atari with deep reinforcement learning.
\newblock \emph{arXiv preprint arXiv:1312.5602}, 2013.

\bibitem[Mnih et~al.(2016)Mnih, Badia, Mirza, Graves, Lillicrap, Harley,
  Silver, and Kavukcuoglu]{mnih2016asynchronous}
Volodymyr Mnih, Adria~Puigdomenech Badia, Mehdi Mirza, Alex Graves, Timothy
  Lillicrap, Tim Harley, David Silver, and Koray Kavukcuoglu.
\newblock Asynchronous methods for deep reinforcement learning.
\newblock In \emph{International conference on machine learning}, pp.\
  1928--1937, 2016.

\bibitem[Murphy et~al.(2001)Murphy, van~der Laan, Robins, and
  Group]{murphy2001marginal}
Susan~A Murphy, Mark~J van~der Laan, James~M Robins, and Conduct Problems
  Prevention~Research Group.
\newblock Marginal mean models for dynamic regimes.
\newblock \emph{Journal of the American Statistical Association}, 96\penalty0
  (456):\penalty0 1410--1423, 2001.

\bibitem[Nachum et~al.(2017)Nachum, Norouzi, Xu, and
  Schuurmans]{nachum2017trust}
Ofir Nachum, Mohammad Norouzi, Kelvin Xu, and Dale Schuurmans.
\newblock Trust-pcl: An off-policy trust region method for continuous control.
\newblock \emph{arXiv preprint arXiv:1707.01891}, 2017.

\bibitem[Nachum et~al.(2019)Nachum, Chow, Dai, and Li]{nachum2019dualdice}
Ofir Nachum, Yinlam Chow, Bo~Dai, and Lihong Li.
\newblock Dualdice: Behavior-agnostic estimation of discounted stationary
  distribution corrections.
\newblock \emph{arXiv preprint arXiv:1906.04733}, 2019.

\bibitem[Nowozin et~al.(2016)Nowozin, Cseke, and Tomioka]{nowozin2016f}
Sebastian Nowozin, Botond Cseke, and Ryota Tomioka.
\newblock f-gan: Training generative neural samplers using variational
  divergence minimization.
\newblock In \emph{Advances in neural information processing systems}, pp.\
  271--279, 2016.

\bibitem[Peters et~al.(2010)Peters, Mulling, and Altun]{peters2010relative}
Jan Peters, Katharina Mulling, and Yasemin Altun.
\newblock Relative entropy policy search.
\newblock In \emph{Twenty-Fourth AAAI Conference on Artificial Intelligence},
  2010.

\bibitem[Puterman(1990)]{puterman1990markov}
Martin~L Puterman.
\newblock Markov decision processes.
\newblock \emph{Handbooks in operations research and management science},
  2:\penalty0 331--434, 1990.

\bibitem[Schulman et~al.(2015)Schulman, Levine, Abbeel, Jordan, and
  Moritz]{schulman2015trust}
John Schulman, Sergey Levine, Pieter Abbeel, Michael Jordan, and Philipp
  Moritz.
\newblock Trust region policy optimization.
\newblock In \emph{International conference on machine learning}, pp.\
  1889--1897, 2015.

\bibitem[Sutton(1995)]{sutton1995virtues}
Richard~S Sutton.
\newblock On the virtues of linear learning and trajectory distributions.
\newblock In \emph{Proceedings of the Workshop on Value Function Approximation,
  Machine Learning Conference}, pp.\ ~85, 1995.

\bibitem[Todorov et~al.(2012)Todorov, Erez, and Tassa]{todorov2012mujoco}
Emanuel Todorov, Tom Erez, and Yuval Tassa.
\newblock Mujoco: A physics engine for model-based control.
\newblock In \emph{2012 IEEE/RSJ International Conference on Intelligent Robots
  and Systems}, pp.\  5026--5033. IEEE, 2012.

\bibitem[Tsitsiklis \& Van~Roy(1997)Tsitsiklis and
  Van~Roy]{tsitsiklis1997analysis}
John~N Tsitsiklis and Benjamin Van~Roy.
\newblock Analysis of temporal-diffference learning with function
  approximation.
\newblock In \emph{Advances in neural information processing systems}, pp.\
  1075--1081, 1997.

\bibitem[Van~Hasselt et~al.(2018)Van~Hasselt, Doron, Strub, Hessel, Sonnerat,
  and Modayil]{van2018deep}
Hado Van~Hasselt, Yotam Doron, Florian Strub, Matteo Hessel, Nicolas Sonnerat,
  and Joseph Modayil.
\newblock Deep reinforcement learning and the deadly triad.
\newblock \emph{arXiv preprint arXiv:1812.02648}, 2018.

\bibitem[Wu et~al.(2019)Wu, Winston, Kaushik, and Lipton]{wu2019domain}
Yifan Wu, Ezra Winston, Divyansh Kaushik, and Zachary Lipton.
\newblock Domain adaptation with asymmetrically-relaxed distribution alignment.
\newblock In \emph{International Conference on Machine Learning}, pp.\
  6872--6881, 2019.

\end{thebibliography}

\newpage
\appendix
\section{Additional Experiment Results}

\subsection{Additional experiment details}

\paragraph{Dataset collection}
For each environment, we collect five datasets: \{no-noise, eps-0.1, eps-0.3, gauss-0.1, gauss-0.3\} using a partially trained policy $\pi$. Each dataset contains 1 million transitions. Different datasets are collected with different injected noise, corresponding to different levels and strategies of exploration. The specific noise configurations are shown below:
\begin{itemize}
\item \textbf{no-noise} : The dataset is collected by purely executing the partially trained policy $\pi$ without adding noise.
\item \textbf{eps-0.1}: We make an epsilon greedy policy $\pi'$ with 0.1 probability. That is, at each step, $\pi'$ has 0.1 probability to take a uniformly random action, otherwise takes the action sampled from $\pi$. The final dataset is a mixture of three parts: 40\% transitions are collected by $\pi'$, 40\% transitions are collected by purely executing $\pi$, the remaining 20\% are collected by a random walk policy which takes a uniformly random action at every step. This mixture is motivated by that one may only want to perform exploration in only a portion of episodes when deploying a policy.
\item \textbf{eps-0.3}: $\pi'$ is an epsilon greedy policy with 0.3 probability to take a random action. We do the same mixture as in eps-0.1.
\item \textbf{gauss-0.1}: $\pi'$ is taken as adding an independent $\mathcal{N}(0, 0.1^2)$ Gaussian noise to each action sampled from $\pi$. We do the same mixture as in eps-0.1.
\item \textbf{gauss-0.3}: $\pi'$ is taken as adding an independent $\mathcal{N}(0, 0.3^2)$ Gaussian noise to each action sampled from $\pi$. We do the same mixture as in eps-0.1.
\end{itemize}

\paragraph{Hyperparameter search}
As we mentioned in main text, for each variant of $\algname$ and each environment, we do a grid search over policy learning rate and regularization strength. 
For policy learning rate, we search over six values: $\{3\cdot 10^6, 1\cdot 10^5, 3\cdot 10^5, 0.0001, 0.0003, 0.001\}$. 
The regularization strength is controlled differently in different algorithms:
\begin{itemize}
\item In BCQ, we search for the perturbation range $\Phi\in \{0.005, 0.015, 0.05, 0.15, 0.5\}$. $0.05$ is the reported value by its paper \citep{fujimoto2018off}.
\item In BEAR the regularization weight $\alpha$ is adaptively trained with dual gradient ascent based on a divergence constraint $\epsilon$ that is tuned as a hyperparameter. We search for $\epsilon \in \{0.015, 0.05, 0.15, 0.5, 1.5\}$. $0.05$ is the reported value by its paper \citep{kumar2019stabilizing}.
\item When MMD is used with a fixed $\alpha$, we search for $\alpha \in \{3, 10, 30, 100, 300\}$.
\item When KL divergence is used with a fixed $\alpha$ (both KL and KL\_dual), we search for $\alpha \in \{0.1, 0.3, 1.0, 3.0, 10.0\}$.
\item When Wasserstein distance is used with a fixed $\alpha$, we search for $\alpha \in \{0.3, 1.0, 3.0, 10.0, 30.0\}$.
\end{itemize}
in BEAR the regularization weight is adaptively trained with dual gradient ascent based on a divergence constraint $\epsilon$ that is tuned as a hyperparameter;

In the simplest case, the regularization weight $\alpha$ is fixed; 
in BCQ the corresponding tuning is for the perturbation range $\Phi$.
For each of these options, we search over five values (see Appendix for details).
For existing algorithms such as BEAR and BCQ, the reported hyperparameters in their papers \citep{kumar2019stabilizing, fujimoto2018off} are included in this search range,
We select the best hyperparameters according to the average performance over all five datasets.

\paragraph{Implementation details}
All experiments are implemented with Tensorflow and executed on CPUs. For all function approximators, we use fully connected neural networks with RELU activations. For policy networks, we use $\mathrm{tanh}(\mathrm{Gaussian})$ on outputs following BEAR \citep{kumar2019stabilizing}, except for BCQ where we follow their open sourced implementation. For BEAR and BCQ we follow the network sizes as in their papers. For other variants of $\algname$, we shrink the policy networks from $(400, 300)$ to $(200, 200)$ and Q-networks from $(400, 300)$ to $(300, 300)$ for saving computation time without losing performance. Q-function learning rate is always 0.001. As in other deep RL algorithms, we maintain source and target Q-functions with an update rate 0.005 per iteration.  For MMD we use Laplacian kernels with bandwidth reported by \cite{fujimoto2018off}. For divergences in the dual form (both KL\_dual and Wasserstein), we training a $(300, 300)$ fully connected network as the critic in the minimax objective. Gradient penalty (one sided version of the penalty in \cite{gulrajani2017improved} with coefficient 5.0) is applied to both KL and Wasserstein dual training. In each training iteration, the dual critic is updated for 3 steps (which we find better than only 1 step) with learning rate 0.0001. We use Adam for all optimizers. Each agent is trained for 0.5 million steps with batch size 256 (except for BCQ we use 100 according their open sourced implementation). At test time we follow \cite{kumar2019stabilizing} and \cite{fujimoto2018off} by sampling 10 actions from $\pi_\theta$ at each step and take the one with highest learned Q-value.

\newpage
\subsection{Value penalty v.s. policy regularization}

\begin{figure}[h]
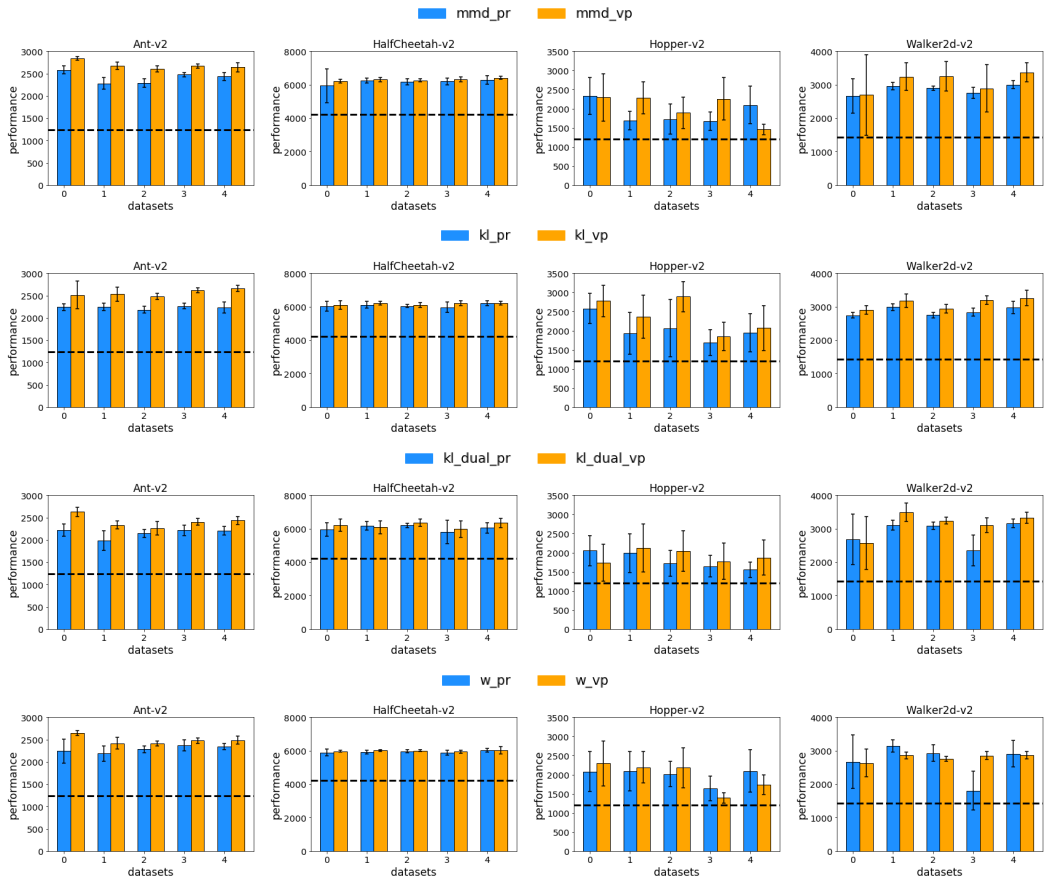

\centering
\includegraphics[height=0.035\columnwidth]{figs_2/pr_vp_mmd_legend.pdf}\\
\includegraphics[width=0.24\columnwidth]{figs_2/pr_vp_mmd_Ant-v2.pdf}
\includegraphics[width=0.24\columnwidth]{figs_2/pr_vp_mmd_HalfCheetah-v2.pdf}
\includegraphics[width=0.24\columnwidth]{figs_2/pr_vp_mmd_Hopper-v2.pdf}
\includegraphics[width=0.24\columnwidth]{figs_2/pr_vp_mmd_Walker2d-v2.pdf}\\
\includegraphics[height=0.035\columnwidth]{figs_2/pr_vp_kl_legend.pdf}\\
\includegraphics[width=0.24\columnwidth]{figs_2/pr_vp_kl_Ant-v2.pdf}
\includegraphics[width=0.24\columnwidth]{figs_2/pr_vp_kl_HalfCheetah-v2.pdf}
\includegraphics[width=0.24\columnwidth]{figs_2/pr_vp_kl_Hopper-v2.pdf}
\includegraphics[width=0.24\columnwidth]{figs_2/pr_vp_kl_Walker2d-v2.pdf}\\
\includegraphics[height=0.035\columnwidth]{figs_2/pr_vp_kl_dual_legend.pdf}\\
\includegraphics[width=0.24\columnwidth]{figs_2/pr_vp_kl_dual_Ant-v2.pdf}
\includegraphics[width=0.24\columnwidth]{figs_2/pr_vp_kl_dual_HalfCheetah-v2.pdf}
\includegraphics[width=0.24\columnwidth]{figs_2/pr_vp_kl_dual_Hopper-v2.pdf}
\includegraphics[width=0.24\columnwidth]{figs_2/pr_vp_kl_dual_Walker2d-v2.pdf}\\
\includegraphics[height=0.035\columnwidth]{figs_2/pr_vp_w_legend.pdf}\\
\includegraphics[width=0.24\columnwidth]{figs_2/pr_vp_w_Ant-v2.pdf}
\includegraphics[width=0.24\columnwidth]{figs_2/pr_vp_w_HalfCheetah-v2.pdf}
\includegraphics[width=0.24\columnwidth]{figs_2/pr_vp_w_Hopper-v2.pdf}
\includegraphics[width=0.24\columnwidth]{figs_2/pr_vp_w_Walker2d-v2.pdf}
\caption{
Comparing policy regularization (pr) v.s. value penalty (vp) with all four divergences. The use of value penalty is usually slightly better.
}
\label{fig:pr_vp_all}
\end{figure}

\newpage
\subsection{Full performance results under different hyperparameters}

\begin{figure}[h]
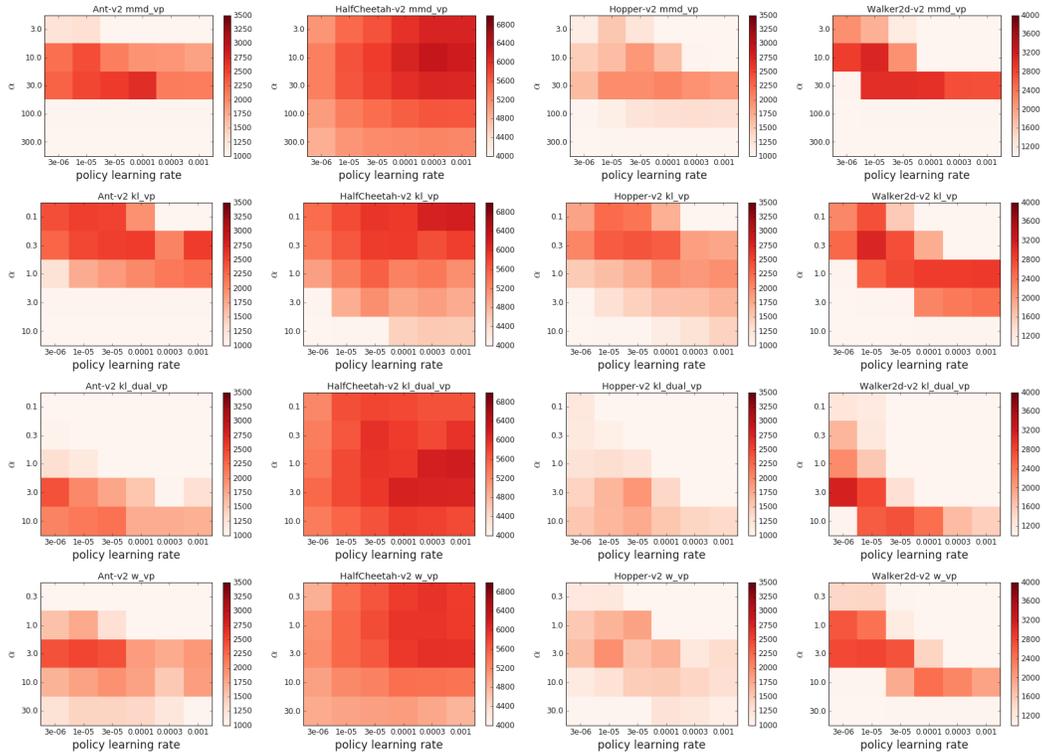

\centering
\includegraphics[width=0.24\textwidth]{figs_grids/bear_vp_Ant-v2.pdf} 
\includegraphics[width=0.24\textwidth]{figs_grids/bear_vp_HalfCheetah-v2.pdf} 
\includegraphics[width=0.24\textwidth]{figs_grids/bear_vp_Hopper-v2.pdf} 
\includegraphics[width=0.24\textwidth]{figs_grids/bear_vp_Walker2d-v2.pdf}
\includegraphics[width=0.24\textwidth]{figs_grids/primal_kl_Ant-v2.pdf} 
\includegraphics[width=0.24\textwidth]{figs_grids/primal_kl_HalfCheetah-v2.pdf} 
\includegraphics[width=0.24\textwidth]{figs_grids/primal_kl_Hopper-v2.pdf} 
\includegraphics[width=0.24\textwidth]{figs_grids/primal_kl_Walker2d-v2.pdf}
\includegraphics[width=0.24\textwidth]{figs_grids/dual_kl_Ant-v2.pdf} 
\includegraphics[width=0.24\textwidth]{figs_grids/dual_kl_HalfCheetah-v2.pdf} 
\includegraphics[width=0.24\textwidth]{figs_grids/dual_kl_Hopper-v2.pdf} 
\includegraphics[width=0.24\textwidth]{figs_grids/dual_kl_Walker2d-v2.pdf}
\includegraphics[width=0.24\textwidth]{figs_grids/dual_w_Ant-v2.pdf} 
\includegraphics[width=0.24\textwidth]{figs_grids/dual_w_HalfCheetah-v2.pdf} 
\includegraphics[width=0.24\textwidth]{figs_grids/dual_w_Hopper-v2.pdf} 
\includegraphics[width=0.24\textwidth]{figs_grids/dual_w_Walker2d-v2.pdf} 
\caption{
Visualization of performance under different hyperparameters. The performance is averaged over all five datasets.
}
\label{fig:grid-search-full}
\end{figure}
\newpage
\begin{figure}[h]
\centering
\includegraphics[width=0.24\textwidth]{figs_grids/bear_Ant-v2.pdf} 
\includegraphics[width=0.24\textwidth]{figs_grids/bear_HalfCheetah-v2.pdf} 
\includegraphics[width=0.24\textwidth]{figs_grids/bear_Hopper-v2.pdf} 
\includegraphics[width=0.24\textwidth]{figs_grids/bear_Walker2d-v2.pdf} 
\includegraphics[width=0.24\textwidth]{figs_grids/primal_kl_novp_Ant-v2.pdf} 
\includegraphics[width=0.24\textwidth]{figs_grids/primal_kl_novp_HalfCheetah-v2.pdf} 
\includegraphics[width=0.24\textwidth]{figs_grids/primal_kl_novp_Hopper-v2.pdf} 
\includegraphics[width=0.24\textwidth]{figs_grids/primal_kl_novp_Walker2d-v2.pdf}
\includegraphics[width=0.24\textwidth]{figs_grids/dual_kl_novp_Ant-v2.pdf} 
\includegraphics[width=0.24\textwidth]{figs_grids/dual_kl_novp_HalfCheetah-v2.pdf} 
\includegraphics[width=0.24\textwidth]{figs_grids/dual_kl_novp_Hopper-v2.pdf} 
\includegraphics[width=0.24\textwidth]{figs_grids/dual_kl_novp_Walker2d-v2.pdf}
\includegraphics[width=0.24\textwidth]{figs_grids/dual_w_novp_Ant-v2.pdf}
\includegraphics[width=0.24\textwidth]{figs_grids/dual_w_novp_HalfCheetah-v2.pdf} 
\includegraphics[width=0.24\textwidth]{figs_grids/dual_w_novp_Hopper-v2.pdf}
\includegraphics[width=0.24\textwidth]{figs_grids/dual_w_novp_Walker2d-v2.pdf}
\caption{
Visualization of performance under different hyperparameters. The performance is averaged over all five datasets.
}
\label{fig:grid-search-full-2}
\end{figure}

\newpage
\begin{figure}[h]
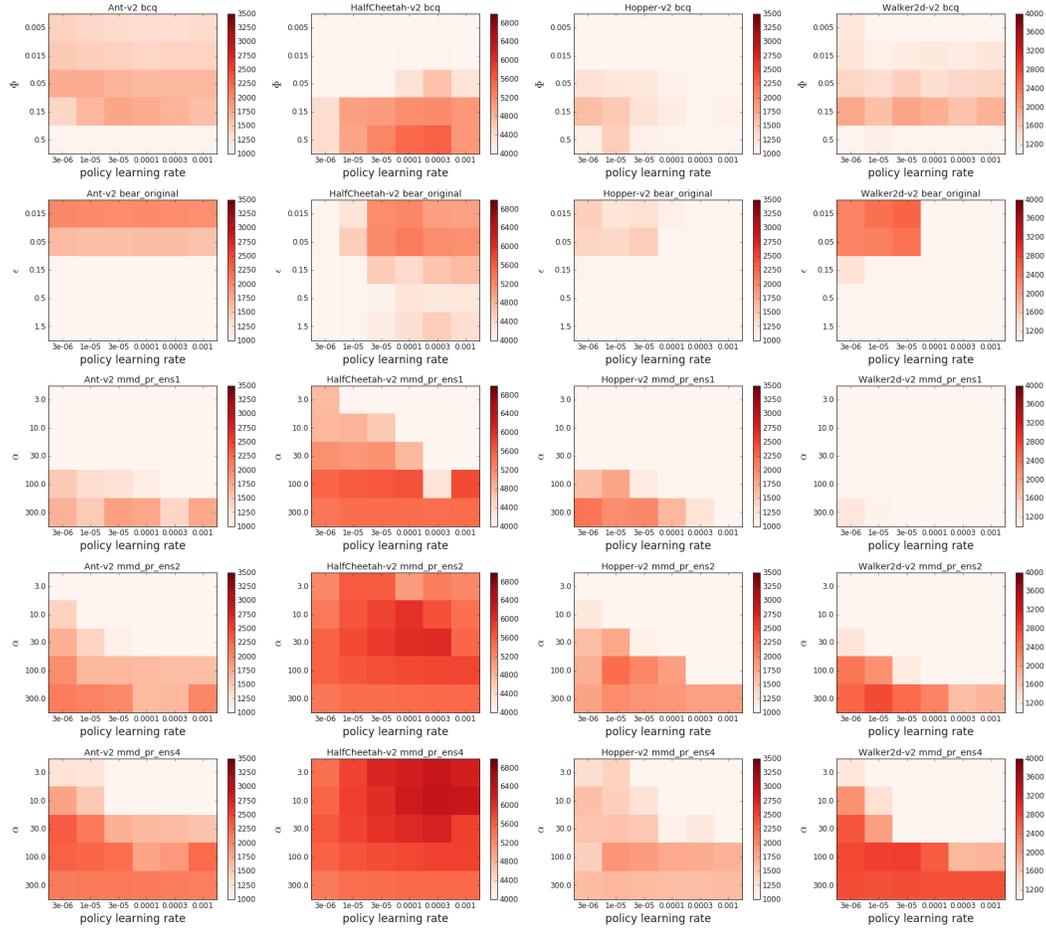

\centering
\includegraphics[width=0.24\textwidth]{figs_grids/bcqori_Ant-v2.pdf} 
\includegraphics[width=0.24\textwidth]{figs_grids/bcqori_HalfCheetah-v2.pdf} 
\includegraphics[width=0.24\textwidth]{figs_grids/bcqori_Hopper-v2.pdf} 
\includegraphics[width=0.24\textwidth]{figs_grids/bcqori_Walker2d-v2.pdf}
\includegraphics[width=0.24\textwidth]{figs_grids/bear_ori_Ant-v2.pdf} 
\includegraphics[width=0.24\textwidth]{figs_grids/bear_ori_HalfCheetah-v2.pdf} 
\includegraphics[width=0.24\textwidth]{figs_grids/bear_ori_Hopper-v2.pdf} 
\includegraphics[width=0.24\textwidth]{figs_grids/bear_ori_Walker2d-v2.pdf}
\includegraphics[width=0.24\textwidth]{figs_grids/bear_ens1_Ant-v2.pdf} 
\includegraphics[width=0.24\textwidth]{figs_grids/bear_ens1_HalfCheetah-v2.pdf} 
\includegraphics[width=0.24\textwidth]{figs_grids/bear_ens1_Hopper-v2.pdf} 
\includegraphics[width=0.24\textwidth]{figs_grids/bear_ens1_Walker2d-v2.pdf}
\includegraphics[width=0.24\textwidth]{figs_grids/bear_ens2_Ant-v2.pdf} 
\includegraphics[width=0.24\textwidth]{figs_grids/bear_ens2_HalfCheetah-v2.pdf} 
\includegraphics[width=0.24\textwidth]{figs_grids/bear_ens2_Hopper-v2.pdf} 
\includegraphics[width=0.24\textwidth]{figs_grids/bear_ens2_Walker2d-v2.pdf}
\includegraphics[width=0.24\textwidth]{figs_grids/bear_ens4_Ant-v2.pdf} 
\includegraphics[width=0.24\textwidth]{figs_grids/bear_ens4_HalfCheetah-v2.pdf} 
\includegraphics[width=0.24\textwidth]{figs_grids/bear_ens4_Hopper-v2.pdf} 
\includegraphics[width=0.24\textwidth]{figs_grids/bear_ens4_Walker2d-v2.pdf}
\caption{
Visualization of performance under different hyperparameters.
}
\label{fig:grid-search-baselines}
\end{figure}

\newpage
\subsection{Additional training curves}

\begin{figure}[h]
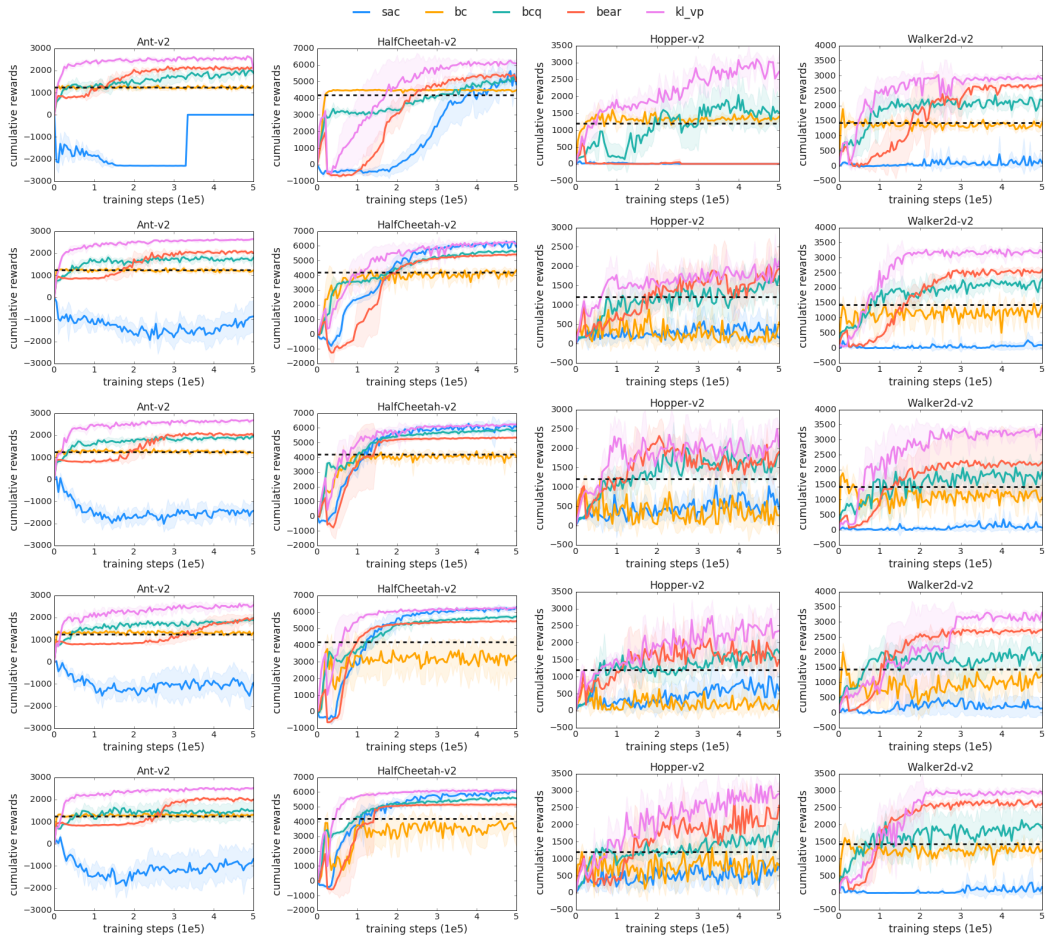

\centering
\includegraphics[width=0.35\columnwidth]{figs_curves/bcq_legend.pdf}\\
\includegraphics[width=0.24\columnwidth]{figs_curves/bcq_Ant-v2_pure.pdf}
\includegraphics[width=0.24\columnwidth]{figs_curves/bcq_HalfCheetah-v2_pure.pdf}
\includegraphics[width=0.24\columnwidth]{figs_curves/bcq_Hopper-v2_pure.pdf}
\includegraphics[width=0.24\columnwidth]{figs_curves/bcq_Walker2d-v2_pure.pdf}
\includegraphics[width=0.24\columnwidth]{figs_curves/bcq_Ant-v2_gaussian1.pdf}
\includegraphics[width=0.24\columnwidth]{figs_curves/bcq_HalfCheetah-v2_gaussian1.pdf}
\includegraphics[width=0.24\columnwidth]{figs_curves/bcq_Hopper-v2_gaussian1.pdf}
\includegraphics[width=0.24\columnwidth]{figs_curves/bcq_Walker2d-v2_gaussian1.pdf}
\includegraphics[width=0.24\columnwidth]{figs_curves/bcq_Ant-v2_gaussian3.pdf}
\includegraphics[width=0.24\columnwidth]{figs_curves/bcq_HalfCheetah-v2_gaussian3.pdf}
\includegraphics[width=0.24\columnwidth]{figs_curves/bcq_Hopper-v2_gaussian3.pdf}
\includegraphics[width=0.24\columnwidth]{figs_curves/bcq_Walker2d-v2_gaussian3.pdf}
\includegraphics[width=0.24\columnwidth]{figs_curves/bcq_Ant-v2_eps1.pdf}
\includegraphics[width=0.24\columnwidth]{figs_curves/bcq_HalfCheetah-v2_eps1.pdf}
\includegraphics[width=0.24\columnwidth]{figs_curves/bcq_Hopper-v2_eps1.pdf}
\includegraphics[width=0.24\columnwidth]{figs_curves/bcq_Walker2d-v2_eps1.pdf}
\includegraphics[width=0.24\columnwidth]{figs_curves/bcq_Ant-v2_eps3.pdf}
\includegraphics[width=0.24\columnwidth]{figs_curves/bcq_HalfCheetah-v2_eps3.pdf}
\includegraphics[width=0.24\columnwidth]{figs_curves/bcq_Hopper-v2_eps3.pdf}
\includegraphics[width=0.24\columnwidth]{figs_curves/bcq_Walker2d-v2_eps3.pdf}
\caption{
Training curves on all five datasets when comparing kl\_vp to other baselines. 
}
\label{fig:curves}
\end{figure}

\newpage

\begin{figure}[h]
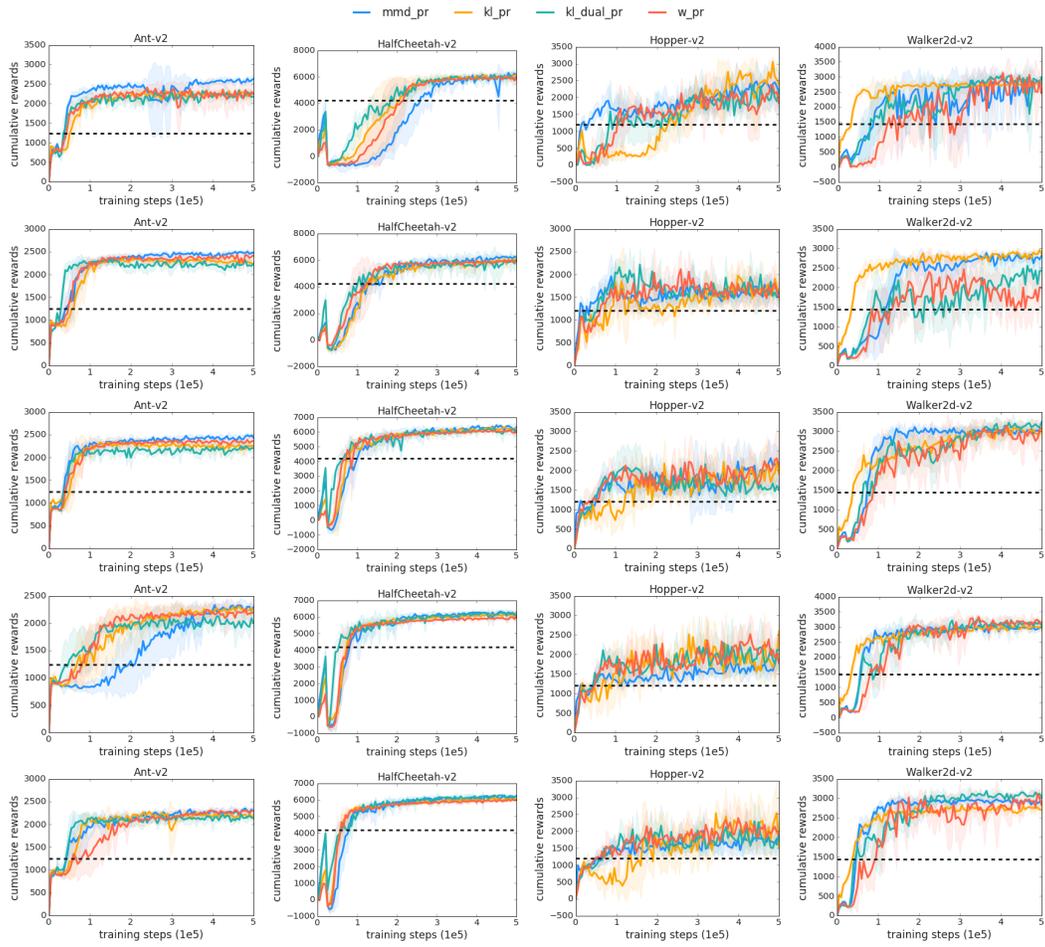

\centering
\includegraphics[width=0.35\columnwidth]{figs_curves/div_pr_legend.pdf}\\
\includegraphics[width=0.24\columnwidth]{figs_curves/div_pr_Ant-v2_pure.pdf}
\includegraphics[width=0.24\columnwidth]{figs_curves/div_pr_HalfCheetah-v2_pure.pdf}
\includegraphics[width=0.24\columnwidth]{figs_curves/div_pr_Hopper-v2_pure.pdf}
\includegraphics[width=0.24\columnwidth]{figs_curves/div_pr_Walker2d-v2_pure.pdf}
\includegraphics[width=0.24\columnwidth]{figs_curves/div_pr_Ant-v2_gaussian1.pdf}
\includegraphics[width=0.24\columnwidth]{figs_curves/div_pr_HalfCheetah-v2_gaussian1.pdf}
\includegraphics[width=0.24\columnwidth]{figs_curves/div_pr_Hopper-v2_gaussian1.pdf}
\includegraphics[width=0.24\columnwidth]{figs_curves/div_pr_Walker2d-v2_gaussian1.pdf}
\includegraphics[width=0.24\columnwidth]{figs_curves/div_pr_Ant-v2_gaussian3.pdf}
\includegraphics[width=0.24\columnwidth]{figs_curves/div_pr_HalfCheetah-v2_gaussian3.pdf}
\includegraphics[width=0.24\columnwidth]{figs_curves/div_pr_Hopper-v2_gaussian3.pdf}
\includegraphics[width=0.24\columnwidth]{figs_curves/div_pr_Walker2d-v2_gaussian3.pdf}
\includegraphics[width=0.24\columnwidth]{figs_curves/div_pr_Ant-v2_eps1.pdf}
\includegraphics[width=0.24\columnwidth]{figs_curves/div_pr_HalfCheetah-v2_eps1.pdf}
\includegraphics[width=0.24\columnwidth]{figs_curves/div_pr_Hopper-v2_eps1.pdf}
\includegraphics[width=0.24\columnwidth]{figs_curves/div_pr_Walker2d-v2_eps1.pdf}
\includegraphics[width=0.24\columnwidth]{figs_curves/div_pr_Ant-v2_eps3.pdf}
\includegraphics[width=0.24\columnwidth]{figs_curves/div_pr_HalfCheetah-v2_eps3.pdf}
\includegraphics[width=0.24\columnwidth]{figs_curves/div_pr_Hopper-v2_eps3.pdf}
\includegraphics[width=0.24\columnwidth]{figs_curves/div_pr_Walker2d-v2_eps3.pdf}
\caption{
Training curves when comparing different divergences with policy regularization. All divergences perform similarly. 
}
\label{fig:div_pr}
\end{figure}

\newpage

\begin{figure}[h]
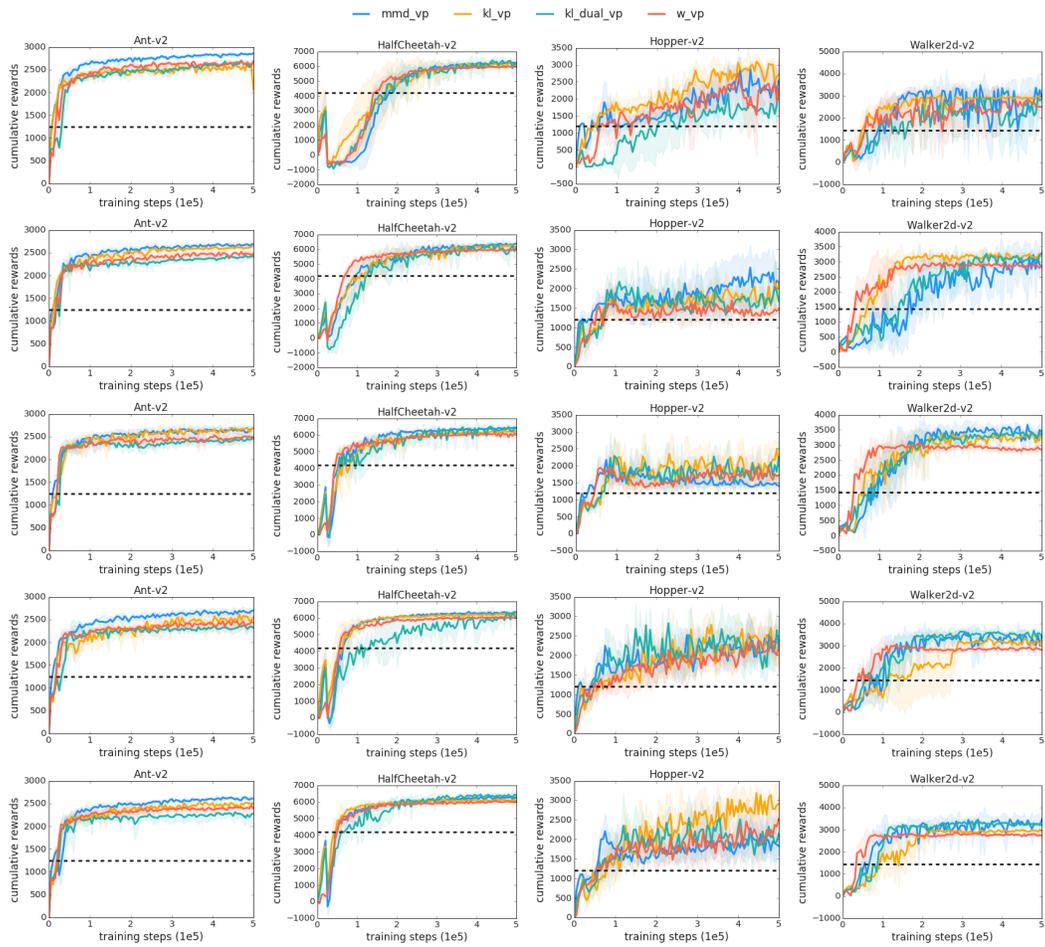

\centering
\includegraphics[width=0.35\columnwidth]{figs_curves/div_vp_legend.pdf}\\
\includegraphics[width=0.24\columnwidth]{figs_curves/div_vp_Ant-v2_pure.pdf}
\includegraphics[width=0.24\columnwidth]{figs_curves/div_vp_HalfCheetah-v2_pure.pdf}
\includegraphics[width=0.24\columnwidth]{figs_curves/div_vp_Hopper-v2_pure.pdf}
\includegraphics[width=0.24\columnwidth]{figs_curves/div_vp_Walker2d-v2_pure.pdf}
\includegraphics[width=0.24\columnwidth]{figs_curves/div_vp_Ant-v2_gaussian1.pdf}
\includegraphics[width=0.24\columnwidth]{figs_curves/div_vp_HalfCheetah-v2_gaussian1.pdf}
\includegraphics[width=0.24\columnwidth]{figs_curves/div_vp_Hopper-v2_gaussian1.pdf}
\includegraphics[width=0.24\columnwidth]{figs_curves/div_vp_Walker2d-v2_gaussian1.pdf}
\includegraphics[width=0.24\columnwidth]{figs_curves/div_vp_Ant-v2_gaussian3.pdf}
\includegraphics[width=0.24\columnwidth]{figs_curves/div_vp_HalfCheetah-v2_gaussian3.pdf}
\includegraphics[width=0.24\columnwidth]{figs_curves/div_vp_Hopper-v2_gaussian3.pdf}
\includegraphics[width=0.24\columnwidth]{figs_curves/div_vp_Walker2d-v2_gaussian3.pdf}
\includegraphics[width=0.24\columnwidth]{figs_curves/div_vp_Ant-v2_eps1.pdf}
\includegraphics[width=0.24\columnwidth]{figs_curves/div_vp_HalfCheetah-v2_eps1.pdf}
\includegraphics[width=0.24\columnwidth]{figs_curves/div_vp_Hopper-v2_eps1.pdf}
\includegraphics[width=0.24\columnwidth]{figs_curves/div_vp_Walker2d-v2_eps1.pdf}
\includegraphics[width=0.24\columnwidth]{figs_curves/div_vp_Ant-v2_eps3.pdf}
\includegraphics[width=0.24\columnwidth]{figs_curves/div_vp_HalfCheetah-v2_eps3.pdf}
\includegraphics[width=0.24\columnwidth]{figs_curves/div_vp_Hopper-v2_eps3.pdf}
\includegraphics[width=0.24\columnwidth]{figs_curves/div_vp_Walker2d-v2_eps3.pdf}
\caption{
Training curves when comparing different divergences with value penalty. All divergences perform similarly.  
}
\label{fig:div_vp}
\end{figure}

\newpage
\subsection{Full performance results under the best hyperparameters} 

\begin{table}[h]
\centering
\begin{tabular}{lccccr}
\toprule
\multicolumn{3}{c}{\bf Environment: Ant-v2} & \multicolumn{3}{c}{\bf Partially trained policy: 1241} \\
\midrule
dataset & no-noise & eps-0.1 & eps-0.3 & gauss-0.1 & gauss-0.3
\\
\midrule
SAC & 0 & -1109 & -911 & -1071 & -1498\\
BC & 1235 & 1300 & 1278 & 1203 & 1240\\
BCQ & 1921 & 1864 & 1504 & 1731 & 1887 \\
BEAR & 2100 & 1897 & 2008 & 2054 & 2018 \\
MMD\_vp      & \textbf{2839}  & \textbf{2672}  & \textbf{2602} & \textbf{2667} & 2640 \\
KL\_vp       & 2514  & 2530  & 2484  & 2615 & \textbf{2661} \\
KL\_dual\_vp &2626 & 2334 & 2256 & 2404 & 2433 \\
W\_vp       & 2646  & 2417  & 2409 & 2474 & 2487 \\
MMD\_pr      & 2583  & 2280  & 2285 & 2477 & 2435 \\
KL\_pr       & 2241  & 2247  & 2181  & 2263 & 2233 \\
KL\_dual\_pr &2218 & 1984 & 2144 & 2215 & 2201 \\
W\_pr       & 2241  & 2186  & 2284  & 2365 & 2344 \\
\bottomrule
\end{tabular}\\
\caption{Evaluation results with tuned hyperparameters. 0 performance means overflow encountered during training due to diverging Q-functions.}
\vspace{-0.1in}
\label{table:full-results-ant}
\end{table}
\begin{table}[h]
\centering
\begin{tabular}{lccccr}
\toprule
\multicolumn{3}{c}{\bf Environment: HalfCheetah-v2} & \multicolumn{3}{c}{\bf Partially trained policy: 4206} \\
\midrule
dataset & no-noise & eps-0.1 & eps-0.3 & gauss-0.1 & gauss-0.3
\\
\midrule
SAC &5093 & 6174 & 5978 & 6082 & 6090 \\
BC &4465 & 3206 & 3751 & 4084 & 4033 \\
BCQ &5064 & 5693 & 5588 & 5614 & 5837 \\
BEAR &5325 & 5435 & 5149 & 5394 & 5329 \\
MMD\_vp      & \textbf{6207} & \textbf{6307} & \textbf{6263} & \textbf{6323} & \textbf{6400} \\
KL\_vp       & 6104 & 6212 & 6104 & 6219 & 6206 \\
KL\_dual\_vp &6209 & 6087 & 6359 & 5972 & 6340 \\
W\_vp       & 5957  & 6014  & 6001  & 5939 & 6025 \\
MMD\_pr      & 5936 & 6242 & 6166 & 6200 & 6294 \\
KL\_pr       & 6032  & 6116  & 6035  & 5969 & 6219 \\
KL\_dual\_pr &5944 & 6183 & 6207 & 5789 & 6050 \\
W\_pr       & 5897 & 5923 & 5970 & 5894 & 6031 \\
\bottomrule
\end{tabular}\\
\caption{Evaluation results with tuned hyperparameters.}
\vspace{-0.1in}
\label{table:full-results-cheetah}
\end{table}

\newpage
\begin{table}[h]
\centering
\begin{tabular}{lccccr}
\toprule
\multicolumn{3}{c}{\bf Environment: Hopper-v2} & \multicolumn{3}{c}{\bf Partially trained policy: 1202} \\
\midrule
dataset & no-noise & eps-0.1 & eps-0.3 & gauss-0.1 & gauss-0.3
\\
\midrule
SAC &0.2655 & 661.7 & 701 & 311.2 & 592.6 \\
BC &1330 & 129.4 & 828.3 & 221.1 & 284.6 \\
BCQ &1543 & 1652 & 1632 & 1599 & 1590 \\
BEAR &0 & 1620 & 2213 & 1825 & 1720 \\
MMD\_vp      & 2291 & 2282 & 1892 & \textbf{2255} & 1458 \\
KL\_vp       & \textbf{2774} & \textbf{2360} & \textbf{2892} & 1851 & 2066 \\
KL\_dual\_vp &1735 & 2121 & 2043 & 1770 & 1872 \\
W\_vp       & 2292  & 2187  & 2178  & 1390 & 1739 \\
MMD\_pr      & 2334 & 1688 & 1725 & 1666 & \textbf{2097} \\
KL\_pr       & 2574  & 1925  & 2064  & 1688 & 1947 \\
KL\_dual\_pr &2053 & 1985 & 1719 & 1641 & 1551 \\
W\_pr       & 2080  & 2089  & 2015  & 1635 & \textbf{2097} \\
\bottomrule
\end{tabular}\\
\caption{Evaluation results with tuned hyperparameters.}
\vspace{-0.1in}
\label{table:full-results-hopper}
\end{table}
\begin{table}[h]
\centering
\begin{tabular}{lccccr}
\toprule
\multicolumn{3}{c}{\bf Environment: Walker-v2} & \multicolumn{3}{c}{\bf Partially trained policy: 1439} \\
\midrule
dataset & no-noise & eps-0.1 & eps-0.3 & gauss-0.1 & gauss-0.3
\\
\midrule
SAC &131.7 & 213.5 & 127.1 & 119.3 & 109.3 \\
BC &1334 & 1092 & 1263 & 1199 & 1137 \\
BCQ &2095 & 1921 & 1953 & 2094 & 1734 \\
BEAR &2646 & 2695 & 2608 & 2539 & 2194 \\
MMD\_vp      & 2694  & 3241  & \textbf{3255} & 2893 & \textbf{3368} \\
KL\_vp       & \textbf{2907}  & 3175  & 2942 & \textbf{3193} & 3261 \\
KL\_dual\_vp & 2575 & \textbf{3490} & 3236 & 3103 & 3333 \\
W\_vp       & 2635  & 2863  & 2758  & 2856 & 2862 \\
MMD\_pr      & 2670  & 2957 & 2897 & 2759 & 3004 \\
KL\_pr       & 2744  & 2990  & 2747  & 2837 & 2981 \\
KL\_dual\_pr & 2682 & 3109 & 3080 & 2357 & 3155 \\
W\_pr       & 2667  & 3140  & 2928  & 1804 & 2907 \\
\bottomrule
\end{tabular}\\
\caption{Evaluation results with tuned hyperparameters.}
\vspace{-0.1in}
\label{table:full-results-walker}
\end{table}

\end{document}